\documentclass[]{spie}  

\usepackage{amsmath,amsfonts,amssymb}
\usepackage{graphicx}
\usepackage[colorlinks=true, allcolors=blue]{hyperref}
\usepackage{booktabs}
\usepackage{multirow}
\usepackage{caption}
\usepackage{subcaption}

\usepackage[]{xcolor}
\definecolor{emHotPink}{rgb}{1.0, 0.0, 0.5}
\definecolor{jioBlue}{rgb}{0.0, 0.0, 1.0}
\definecolor{emBlack}{rgb}{0.0, 0.0, 0.0}

\title{Assessing Coarse-to-Fine Deep~Learning~Models for Optic~Disc~and~Cup Segmentation in Fundus~Images}

\author[a,b]{Eugenia Moris}
\author[a,b]{Nicolás Dazeo}
\author[c]{María~Paula~Albina~de~Rueda}
\author[c]{Francisco~Filizzola}
\author[c]{Nicolás~Iannuzzo}
\author[c]{Danila~Nejamkin}
\author[c]{Kevin~Wignall}
\author[c]{Mercedes~Leguía}
\author[a,b]{Ignacio~Larrabide}
\author[a,b]{José~Ignacio~Orlando}

\affil[a]{Yatiris Group, PLADEMA Institute, UNICEN, Campus Universitario, Tandil, Argentina}
\affil[b]{Consejo Nacional de Investigaciones Cient\'ificas y T\'ecnicas, CONICET, Tandil, Argentina}
\affil[c]{Servicio de Oftalmolog\'ia, Hospital de Alta Complejidad En Red "El Cruce" Dr.~N\'estor~Carlos~Kirchner, Av. Calchaqu\'i 5401, Florencio Varela, Argentina}

\authorinfo{Further author information: (Send correspondence to E.M.)\\E.M..: E-mail: emoris@pladema.exa.unicen.edu.ar, Telephone: +54 249 438-5690\\  }

\begin{document} 
\maketitle

\begin{abstract}
Automated optic disc (OD) and optic cup (OC) segmentation in fundus images is relevant to efficiently measure the vertical cup-to-disc ratio (vCDR), a biomarker commonly used in ophthalmology to determine the degree of glaucomatous optic neuropathy.
In general this is solved using coarse-to-fine deep learning algorithms in which a first stage approximates the OD and a second one uses a crop of this area to predict OD/OC masks.
While this approach is widely applied in the literature, there are no studies analyzing its real contribution to the results.
In this paper we present a comprehensive analysis of different coarse-to-fine designs for OD/OC segmentation using 5 public databases, both from a standard segmentation perspective and for estimating the vCDR for glaucoma assessment.
Our analysis shows that these algorithms not necessarily outperfom standard multi-class single-stage models, especially when these are learned from sufficiently large and diverse training sets.
Furthermore, we noticed that the coarse stage achieves better OD segmentation results than the fine one, and that providing OD supervision to the second stage is essential to ensure accurate OC masks.
Moreover, both the single-stage and two-stage models trained on a multi-dataset setting showed results close to other state-of-the-art alternatives in REFUGE and DRISHTI.
Finally, we evaluated the models for vCDR prediction in comparison with six ophthalmologists on a subset of AIROGS images, to understand them in the context of inter-observer variability.
We noticed that vCDR estimates recovered both from single-stage and coarse-to-fine models can obtain good glaucoma detection results even when they are not highly correlated with manual measurements from experts.
To ensure the reproducibility of our study, our results, our multi-expert dataset and further implementation details are made publicly available at \url{https://github.com/eugeniaMoris/sipaim-2022-coarse-to-fine}.
\end{abstract}


\section{INTRODUCTION}
\label{sec:intro}

Glaucoma is the second leading cause of preventable blindness in the world\cite{orlando2020refuge,kovalyk2022papila}. 
By 2040, it is estimated that 111.8 million people with ages between 40 and 80 years will suffer from this condition worldwide~\cite{kovalyk2022papila}. 
This is attributed to the asymptomatic characteristics of glaucoma, which is produced by an unstable or sustained imperceptible increase in the intra-ocular pressure that silently damages the optic nerve head and causes a gradual yet irreversible loss of vision. 
Hence, half of the people with glaucoma do not know they are affected \cite{phene2019deep}, with this rate rising up to 90\% in some developing countries\cite{phene2019deep}.

Treatments to control disease progression and therefore avoid vision loss heavily rely on early diagnosis \cite{phene2019deep}. 
Color fundus photography is a cost-effective medical imaging modality widely applied for detecting eye diseases such as diabetic retinopathy or age-related macular degeneration~\cite{li2021applications}. 
Early signs of glaucoma involve abnormal changes in the depth of the optic disc (OD) such as optic nerve cupping \cite{phene2019deep}, which are difficult to observe in these projective 2D scans~\cite{orlando2020refuge}.
Nevertheless, as glaucomatous optic neuropathy is characterized by the vertical elongation of the optic cup (OC) \cite{crowston2004effectvCDR}, ophthalmologists frequently use fundus photography to calculate the so-called vertical cup-to-disc ratio (vCDR), a biomarker that aids to quantify the glaucomatous loss of the neuroretinal rim \cite{crowston2004effectvCDR}. 
Measuring this parameter requires to manually determine the vertical diameter of the OD and the OC, which is prone to high inter-observer variability \cite{guo2019automatic}.
Alternatively, several approaches have been proposed to automatically segment these two structures and then compute the vCDR is a much more efficient and accurate way \cite{li2021applications,orlando2020refuge}.

\begin{figure}[t]
    \centering
    
    \begin{subfigure}[]{\textwidth}
      \includegraphics[width=\textwidth]{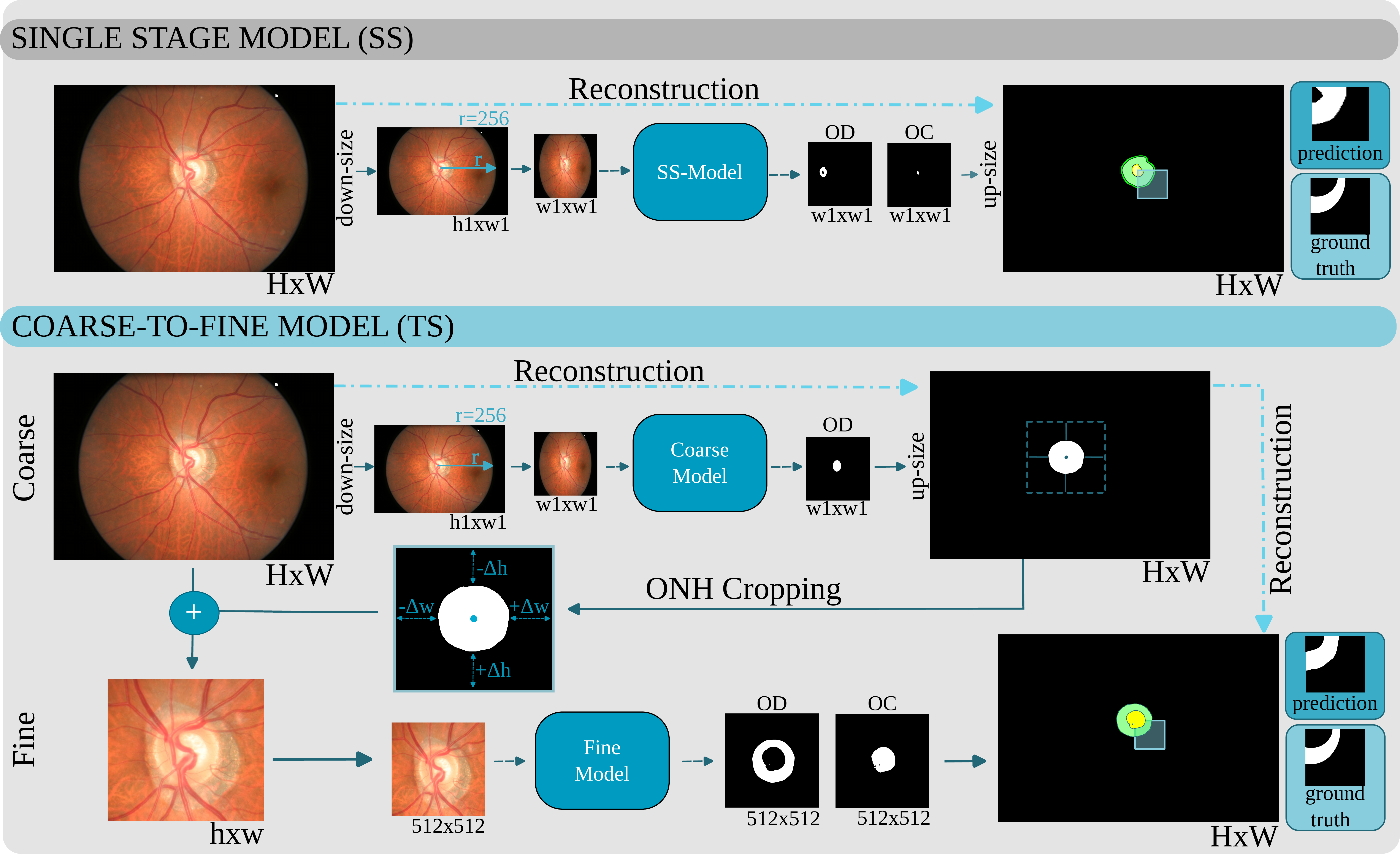}
      \caption{}
      \label{fig:schematic-models}
    \end{subfigure}
    
    \vspace{0.3cm}
    
    \begin{subfigure}[]{\textwidth}
      \includegraphics[width=\textwidth]{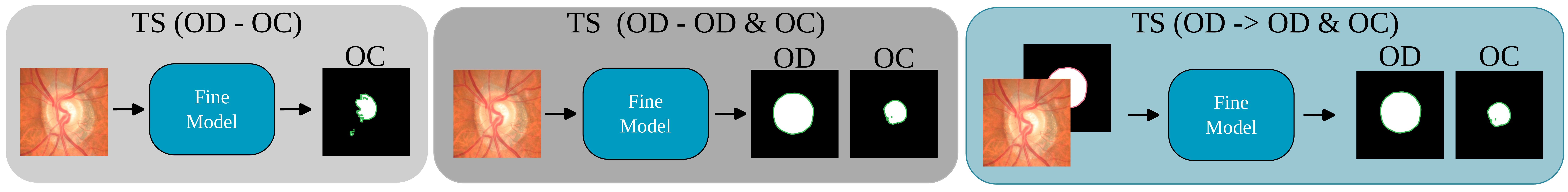}
      \caption{}
      \label{fig:schematic-two-stage}
    \end{subfigure}
    
    \vspace{0.2cm}
    \caption{(a) Schematic representation of single-stage (SS) and coarse-to-fine two-stage (TS) models for OD/OC segmentation in fundus images. SS models segment OD/OC using a single network applied to downsized versions of the original images. TS models, on the other hand, use a coarse stage to detect the OD area, which is cropped and used to feed a fine stage that produces the final OD/OC masks. (b) Different versions of the fine stage considered for analysis: using only OC supervision (left), using both OD and OC supervision (center), and using OD/OC supervision but including the coarse stage OD prediction in the input.}
    \label{fig:schematic}
\end{figure}

Current methods for OD/OC segmentation approach the problem as a multi-class segmentation task using fully convolutional neural networks\cite{liu2021joint,shah2019dynamic,wang2019patch,tabassum2020cded,al2018dense}. 
These models are learned using stochastic gradient descent over mini-batches of images resized to a fixed resolution, which in this case is way much smaller than the original one. 
Hence, several authors have hypothesized that this downsizing process might affect the edges of the target labels and therefore produce sub-optimal results.\cite{al2018dense,wang2019patch,liu2021joint,shah2019dynamic}
To alleviate this issue, multiple existing OD/OC segmentation models rely on coarse-to-fine approaches based on serially aligned architectures, as depicted in Figure~\ref{fig:schematic}. 
In this setting, a first network takes a resized fundus picture as input and predicts a coarse pixel-wise probability map for the OD class. 
The output is then up-sampled to match the original resolution of the image and used for cropping the area around the optic nerve head. The resulting image is then used to feed a second network, which produces a much more detailed segmentation of both the OD and the OC.
This configuration has been used in multiple submissions to REFUGE~\cite{orlando2020refuge} and REFUGE 2~\cite{fang2022refuge2} challenges, with variations usually focused on changing the backbone architectures~\cite{wang2019patch,shah2019dynamic} or replacing the segmentation network from the first stage by an object detector~\cite{liu2021joint}. 
In all these cases, the main hypothesis is that feeding the second stage with an image in which the optic nerve appears approximately at the same resolution as in the original image will result in a much more accurate segmentation.
To the best of our knowledge, however, this two-stage design has always been applied without an empirical in-depth analysis of its contribution to the final results e.g. through an ablation study.
Hence, it is uncertain if this decision holds accurate in general clinical scenarios.

In this paper we present a comprehensive empirical assessment of coarse-to-fine models for OD/OC segmentation in fundus images. 
Instead of studying a single standard design of "first OD, then OD/OC", we tested a series of other two-stage alternatives and compared them all with respect to the more straightforward single-stage multi-class model. 
As far as we know, this is the first in-depth study assessing the utility of these network designs.
Our evaluation using multiple public datasets, including the recently published PAPILA~\cite{kovalyk2022papila}, indicates that coarse-to-fine algorithms not necessarily outperform the single-stage models, especially if these are learned from sufficiently large and variable training sets.
This evaluation also allowed us to draw several conclusions about the relevance of OD information for ensuring the fine stage to produce accurate OC masks.
On the other hand, we complement the study by assessing the effectiveness of the final segmentations for predicting vCDR and using it for glaucoma detection. 
We observed that both single-stage and coarse-to-fine models performed similarly for this task, even when vCDR predictions are not highly correlated with manually produced measurements.
Finally, we also created a dataset with vCDR information produced by six experts on a subset of 40 images from AIROGS,~\footnote{\url{https://airogs.grand-challenge.org/Home/}} that we used to understand results in the context of inter-observer variability.

\section{MATERIALS AND METHODS}

\subsection{Coarse-to-fine two-stage models for OD/OC segmentation}

The purpose of this study is to understand the contribution of coarse-to-fine two-stage (TS) architectural designs over standard multi-class single-stage (SS) models for OD/OC segmentation in fundus images. SS-models are traditional segmentation networks that simultaneously predict score maps for OD/OC areas from input images resized to a pre-defined resolution (Figure~\ref{fig:schematic-models}, top). TS-models, on the other hand, are coarse-to-fine approaches that use a first stage to approximate the location of the OD and a second one to segment both structures at a larger resolution (Figure~\ref{fig:schematic-models}, bottom). To cover as many designs as possible, we studied the following three alternative versions of TS approaches, namely:

\noindent \textbf{TS-model OD - OC:} in which the first stage predicts the OD and the second one only segments the OC (Figure~\ref{fig:schematic-two-stage}, left). This allows the entire capacity of the second model to be dedicated explicitly to segmenting the OC, but without explicit information about the OD. This strategy was followed by several teams in REFUGE challenge~\cite{orlando2020refuge}.

\noindent \textbf{TS-model OD - OD\&OC:} which corresponds to the most widely applied approach~\cite{wang2019patch,shah2019dynamic,liu2021joint} (Figure~\ref{fig:schematic-two-stage}, center), where the first stage classifies only the OD. The resulting segmentation is used to crop the area of interest from the original image, which is then resized to feed a second model that segments both the OD and the OC from it.

\noindent \textbf{TS-model OD $\rightarrow$ OD\&OC:} in which the first stage is also preserved but its cropped output is concatenated with the cropped image channel-wise to feed the second OD/OC stage. We hypothesize that this auxiliary input can simplify the task to the second stage by providing an approximation to the region of interest, which might act as an attention area and therefore allow to obtain better results.

All models were trained using the same preprocessed images. First, an approximate mask of the field-of-view (FOV) was estimated using the Hough Gradient method.
Training and validation scans were all cropped around this area using its bounding box. Subsequently, these images and their labels were downsampled to a fixed resolution of $512 \times 512$ pixels before feeding the network. In test time, input images were not cropped but resized so that the FOV appeared at approximately the same diameter than training and validation scans. Hence, the network could be applied in a fully convolutional manner to the input image regardless of its size, while ensuring at the same time that the overall shape of the optic nerve head matches the one used for training.

Both networks in the TS-models were trained offline one from another. As the fine stage network has no access during training to the predictions of the coarse stage, the OD ground truth segmentations were used online to obtain the cropped versions of the training and validation images. To this end, the central coordinate and the vertical and horizontal diameters of the OD were estimated for each loaded image and used to calculate a bounding box around the area. Margins of size $\Delta = 10\%$ of the box width and height were used at left/right and top/bottom, respectively, to ensure that the cropped region covered the entire optic nerve head and part of its surrounding tissues. To add robustness to future mispredictions of the first stage that might cause the segmented OD to be not always at the center of the image, we added random noise to the center prediction with a 50\% probability and up to 10\% of its original location. Finally, the cropped image and its associated cropped mask were resized to $512 \times 512$ pixels and used as input for the second stage.

\subsection{Materials}

\begin{table}[]
\centering
\caption{Datasets used in our study for training, validation and testing coarse OD segmentation models (first stage of TS-models) and multiclass OD/OC networks (SS-model and the second stage of TS-model)}
\vspace{0.2cm}
\resizebox{0.7\textwidth}{!}{ 
\begin{tabular}{c|ccc|ccc|ccc}
\hline
\multirow{2}{*}{\centering Datasets} & \multicolumn{3}{c}{Data characteristics} & \multicolumn{3}{|c|}{OD segmentation}      & \multicolumn{3}{c}{OD/OC segmentation}   \\ \cline{2-10}
                    & \# img & OD Mask & \multicolumn{1}{c|}{OC Mask} & Train & Valid & \multicolumn{1}{c|}{Test} & Train & Valid & \multicolumn{1}{c}{Test} \\ \hline
\multicolumn{1}{c|}{REFUGE \cite{orlando2020refuge}}     & 1200   & yes     & \multicolumn{1}{c|}{yes}     & 400   & 400   & \multicolumn{1}{c|}{400}  & 400   & 400   & \multicolumn{1}{c}{400}  \\
\multicolumn{1}{c|}{RIGA \cite{almazroa2018retinal}}     & 749    & yes     & \multicolumn{1}{c|}{yes}     & 675   & 74    & \multicolumn{1}{c|}{-}    & 675   & 74    & \multicolumn{1}{c}{-}    \\
\multicolumn{1}{c|}{ORIGA \cite{zhang2010origa}}         & 650    & yes     & \multicolumn{1}{c|}{yes}     & -     & -     & \multicolumn{1}{c|}{650}  & -     & -     & \multicolumn{1}{c}{650}  \\
\multicolumn{1}{c|}{PAPILA \cite{kovalyk2022papila}}     & 489    & yes     & \multicolumn{1}{c|}{yes}     & -     & -     & \multicolumn{1}{c|}{489}  & -     & -     & \multicolumn{1}{c}{489}  \\
\multicolumn{1}{c|}{RIM ONE v3 \cite{fumero2011rim}}     & 159    & yes     & \multicolumn{1}{c|}{yes}     & -     & -     & \multicolumn{1}{c|}{159}  & -     & -     & \multicolumn{1}{c}{159}  \\
\multicolumn{1}{c|}{DRISHTI \cite{sivaswamy2014drishti}} & 100    & yes     & \multicolumn{1}{c|}{yes}     & 45    & 5     & \multicolumn{1}{c|}{50}   & 45    & 5     & \multicolumn{1}{c}{50}   \\
\multicolumn{1}{c|}{IDRID \cite{porwal2018indian}}       & 81     & yes     & \multicolumn{1}{c|}{no}     & 49    & 5     & \multicolumn{1}{c|}{27}   & -     & -     & \multicolumn{1}{c}{-}    \\ \hline
 \multicolumn{1}{c|}{Multi-dataset}  & 3428  & yes  & yes/no & 1169  & 484   & 1775                      & 1120  & 479   & 1748                      \\ \hline
\end{tabular}}
\label{tab:datasets}
\end{table}

Table~\ref{tab:datasets} lists the datasets used to train, validate and test all segmentation models. Except for RIM ONE v3, which comprises cropped images around the optic disc area, all other databases correspond to standard full size retinal images. We used the pre-defined training, validation and test partitions of REFUGE and DRISHTI to study how training data influences the results of different models. Furthermore, we crafted an additional data partition with images from multiple databases, to maximize the diversity of training samples. The final configuration in training, validation and test of these multi-dataset setting is detailed at the bottom of Table~\ref{tab:datasets}. We followed pre-defined partitions when available, and 10\% of the original data was preserved for validation when it was not divided. Notice also that we preserved ORIGA, RIM ONE v3 and PAPILA for testing, to maximize the comparison with other state-of-the-art approaches. All test sets were also used for evaluating the final segmentations in terms of their ability for predicting the vCDR. Furthermore, we created an additional database with 40 images collected from AIROGS (20 with glaucoma and 20 without it) in which 6 experts (MPAR, FF, NI, DN, KW and ML, co-authors of the paper) calculated the vertical diameter of the OD/OC for computing the vCDR. This set allows to study results in the context of inter-observer variability and is publicly released in our repository.

\subsection{Training setup}

We used the same U-Net architecture as backbone for all models due to its popularity for OD/OC segmentation~\cite{alawad2022machine,shah2019dynamic,wang2019patch,liu2021joint}. 
Its encoder consisted of 4 convolutional blocks with increasing output depths of 64, 128, 256 and 512, each followed by $2 \times 2$ max-pooling layers for downsampling. 
These blocks comprised two consecutive convolutional layers with $3 \times 3$ filters, each followed by a batch normalization layer and a ReLU activation. 
An additional block with 1024 filters was used at the bottleneck, followed by the decoder layers. 
This part used 4 upsampling blocks with 512, 256, 128 and 64 filters each, where each block was based on a bilinear upsampling operation followed by a convolutional block. All models were trained using batches of 5 images, with input intensities standardized using each image mean and standard deviation. The same random, online data augmentation strategies were used for all models, consisting of vertical and horizontal flippings, gaussian blurring, brightness, contrast and saturation jitterings and random rotations and scalings. Their parameters and configurations were empirically selected based on the overall performance on a validation set. Adam optimization with an initial learning rate of 0.01 was used for 200 epochs to minimize the cross-entropy loss. When training using only DRISHTI data, 400 epochs were used instead. We also reduced the learning rate by a factor of 0.2 every time the performance plateaued for a maximum of 2 epochs. The best model through epochs according to the validation set was preserved and finally used for evaluation.

\subsection{Evaluation metrics}

Segmentation results were evaluated using Dice index (which is mathematically equivalent to F1-score, and therefore not sensitive to class imbalance). Furthermore, ground truth and predicted OD/OC masks were used to estimate the vCDR, obtained as the ratio between the vertical diameters of the OC and OD. The estimated vCDR was then compared with values obtained from manual segmentation using the $R$ Pearson correlation coefficient and scatter plots. Finally, we evaluated the accuracy of estimated vCDR values for predicting glaucoma using receiver operating characteristic (ROC) curves and the area under the curve (AUC).

\section{RESULTS \& DISCUSSION}



\subsection{OD/OC segmentation analysis}
\label{subsec:results-segmentation}

\begin{figure}[t]
    \centering
    \begin{subfigure}[]{0.49\textwidth}
      \includegraphics[width=\textwidth]{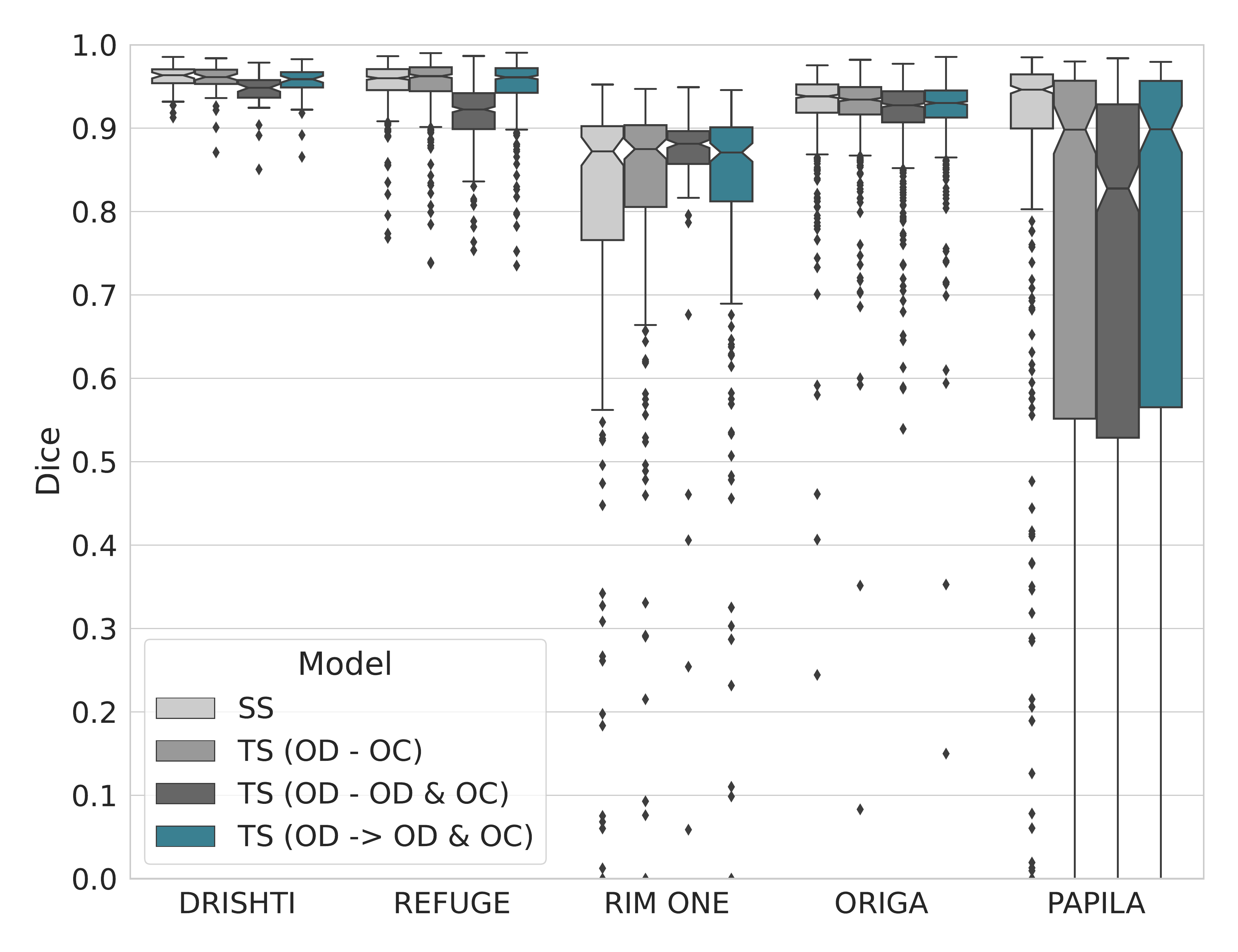}
      \caption{Optic disc (OD) segmentation}
      \label{fig:ablation-study-od}
    \end{subfigure}
    \begin{subfigure}[]{0.49\textwidth}
      \includegraphics[width=\textwidth]{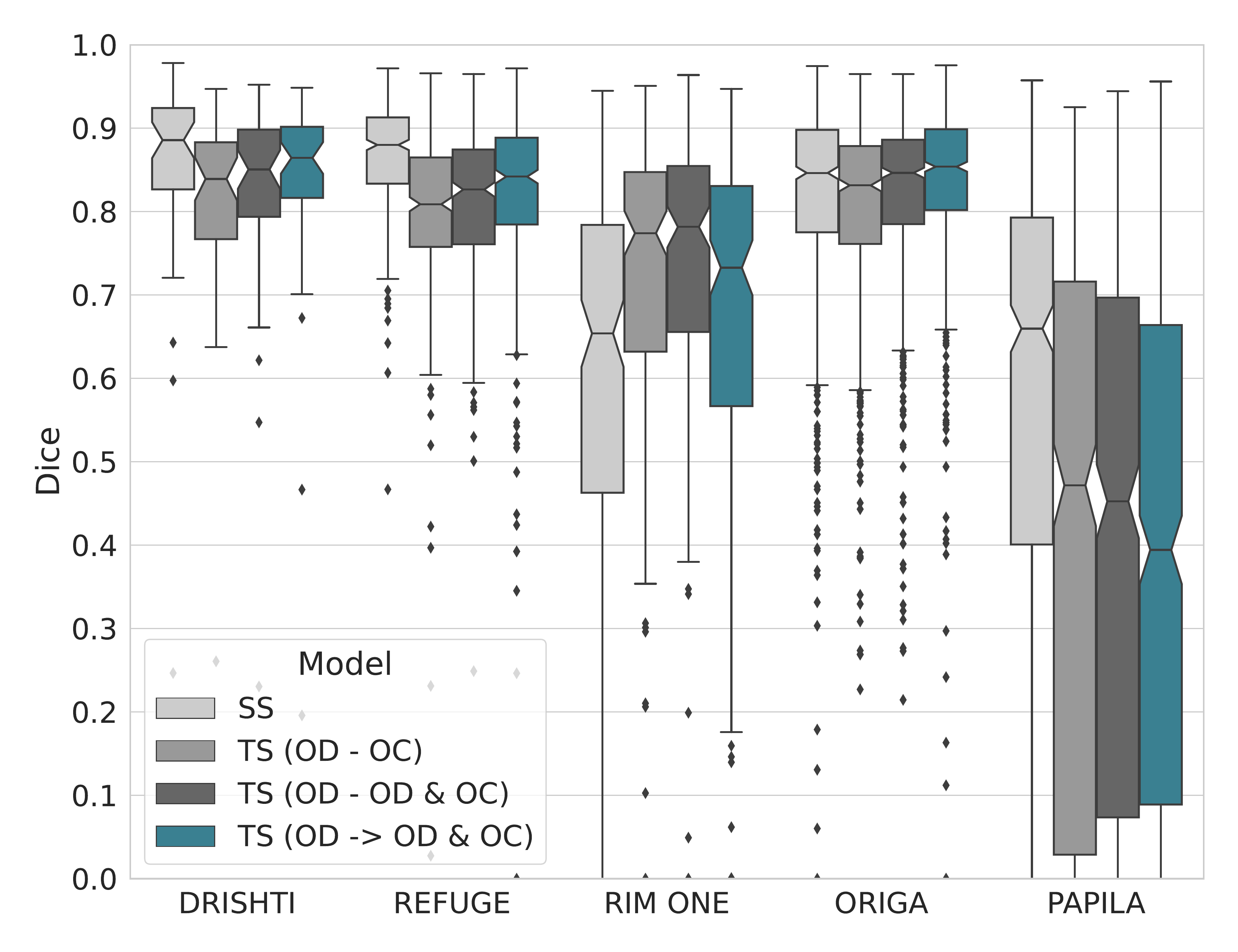}
      \caption{Optic cup (OC) segmentation}
      \label{fig:ablation-study-oc}
    \end{subfigure}
    \vspace{0.2cm}
    \caption{Distribution of Dice values for OD and OC segmentation obtained in each test set by the SS-model and all coarse-to-fine TS-model variants.}
    \label{fig:ablation-study}
\end{figure}

\begin{figure}
    \centering
    \subcaptionbox{REFUGE}{\includegraphics[width=0.24\textwidth]{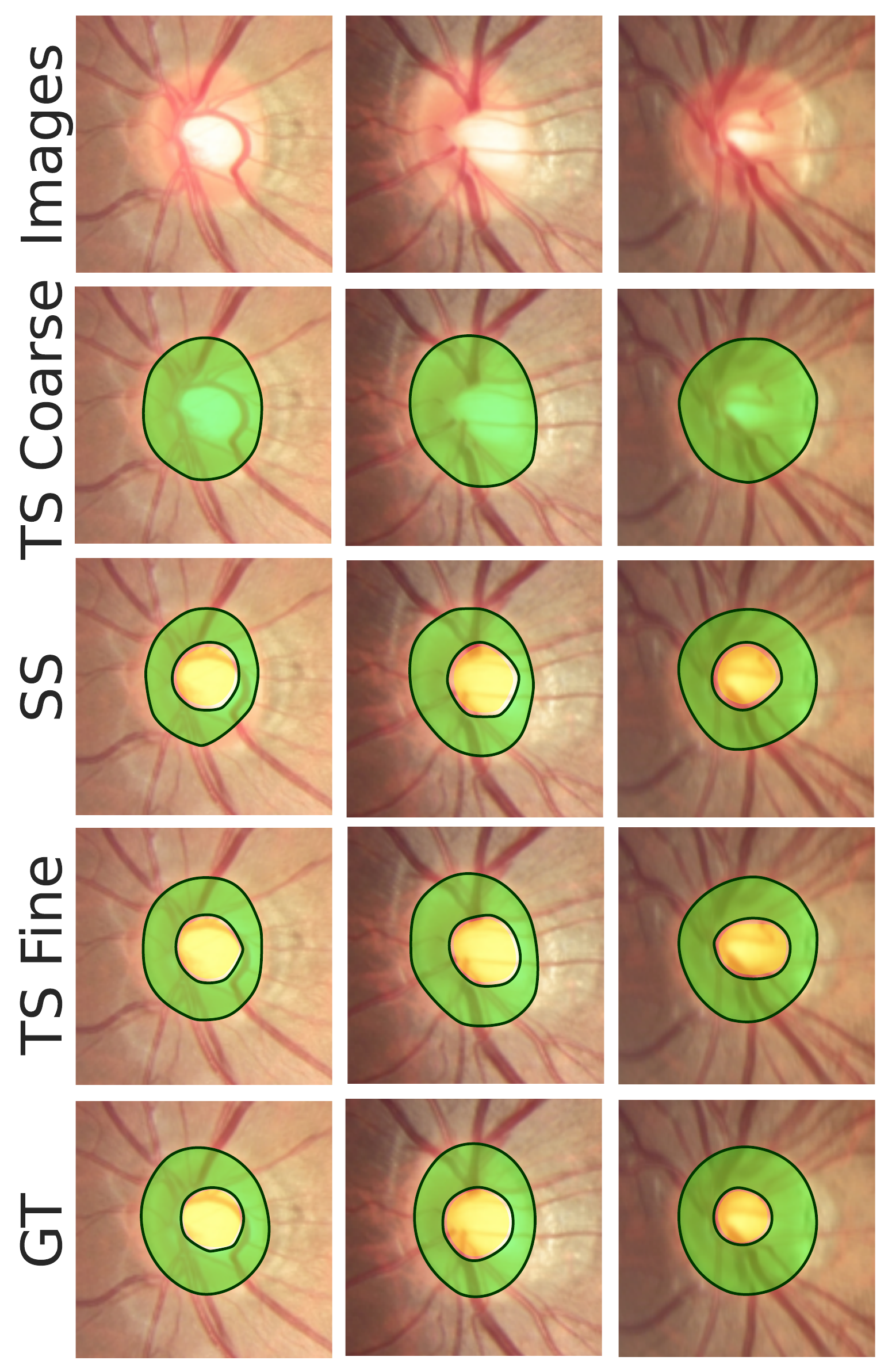}}%
    \hspace{0.001\textwidth}
    \subcaptionbox{DRISHTI}
    {\includegraphics[width=0.24\textwidth]{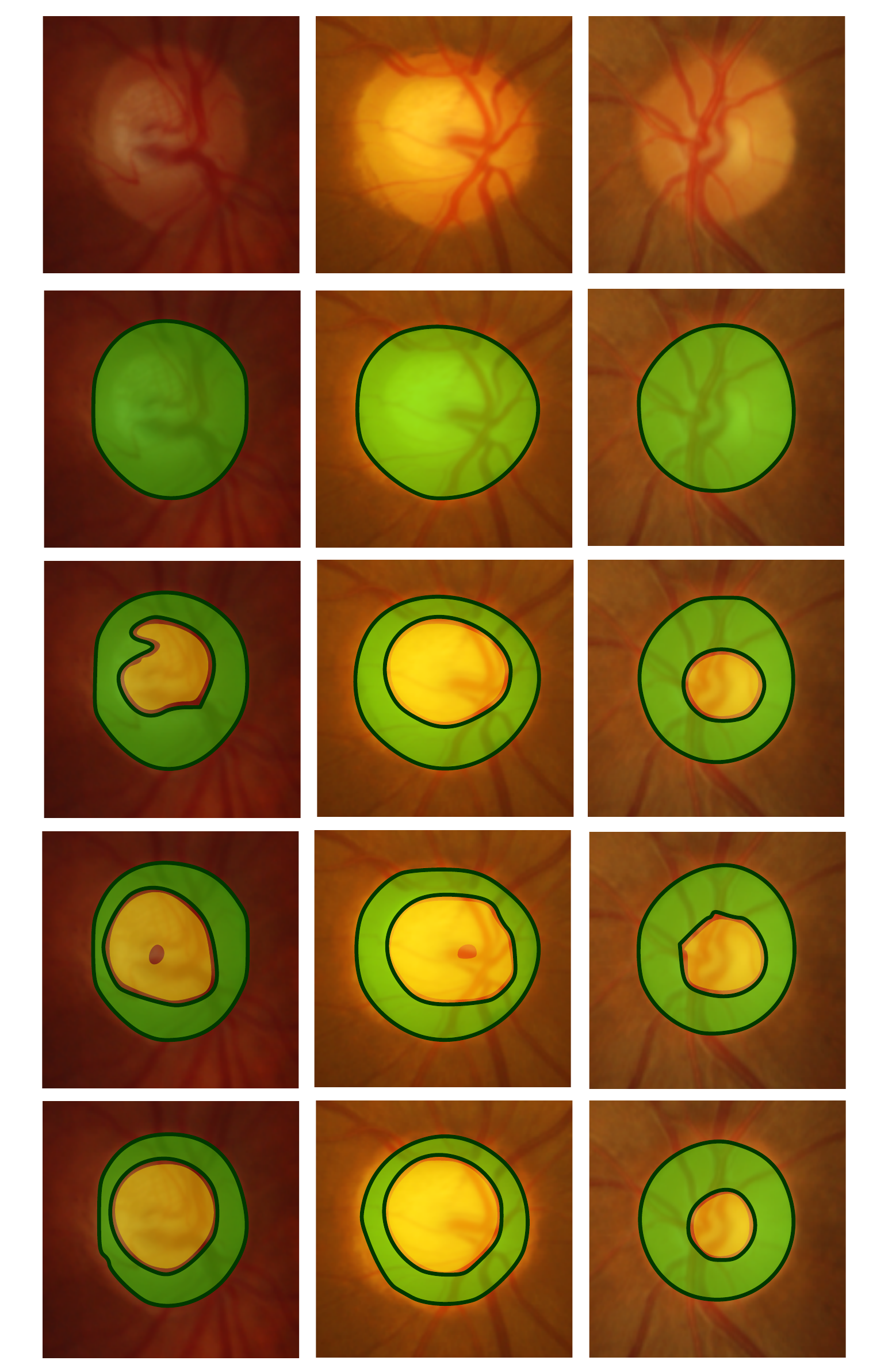}}%
    \subcaptionbox{RIM ONE v3}
    {\includegraphics[width=0.24\textwidth]{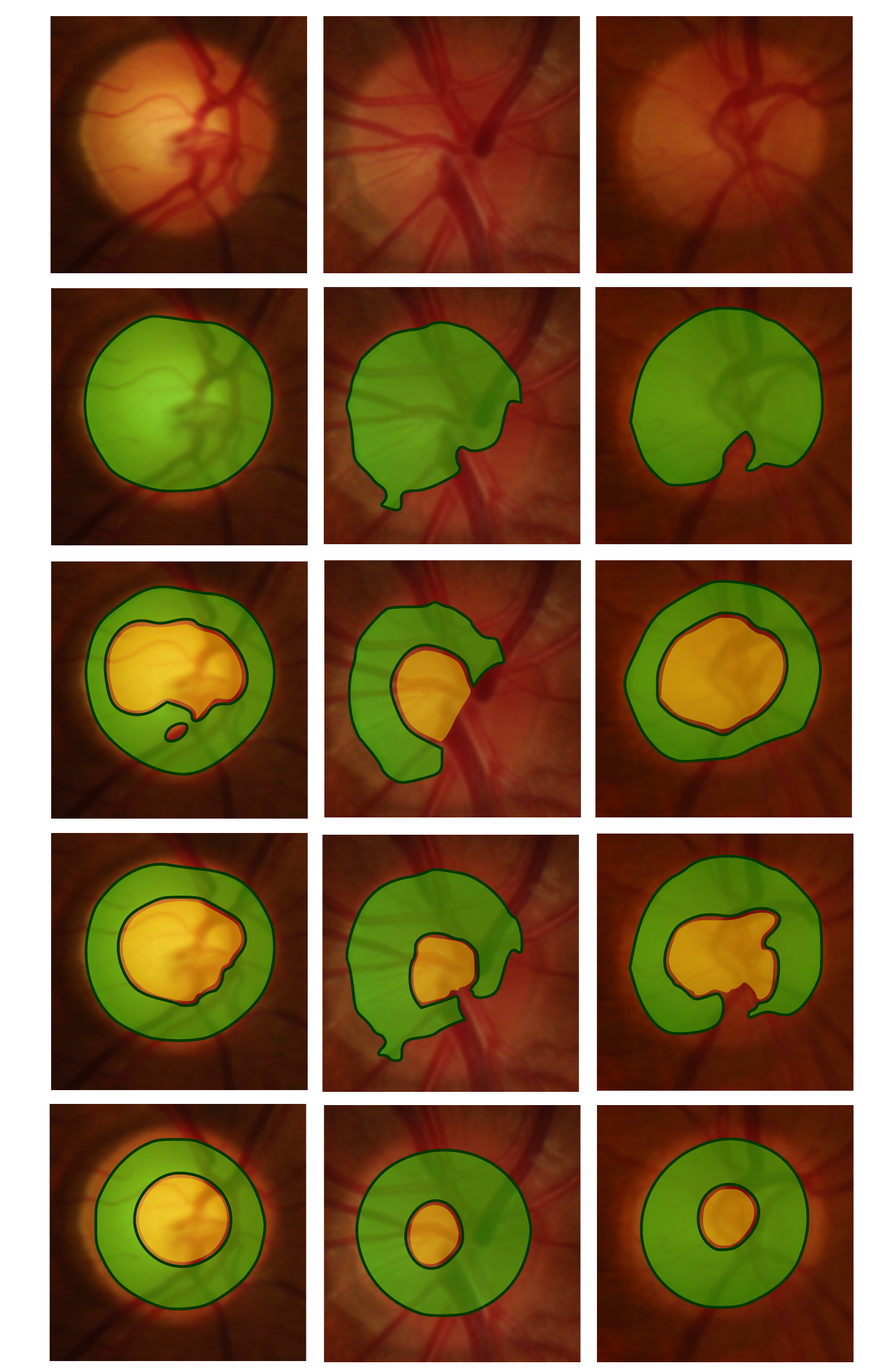}}%
    \hspace{0.001\textwidth}
    \subcaptionbox{ORIGA}
    {\includegraphics[width=0.24\textwidth]{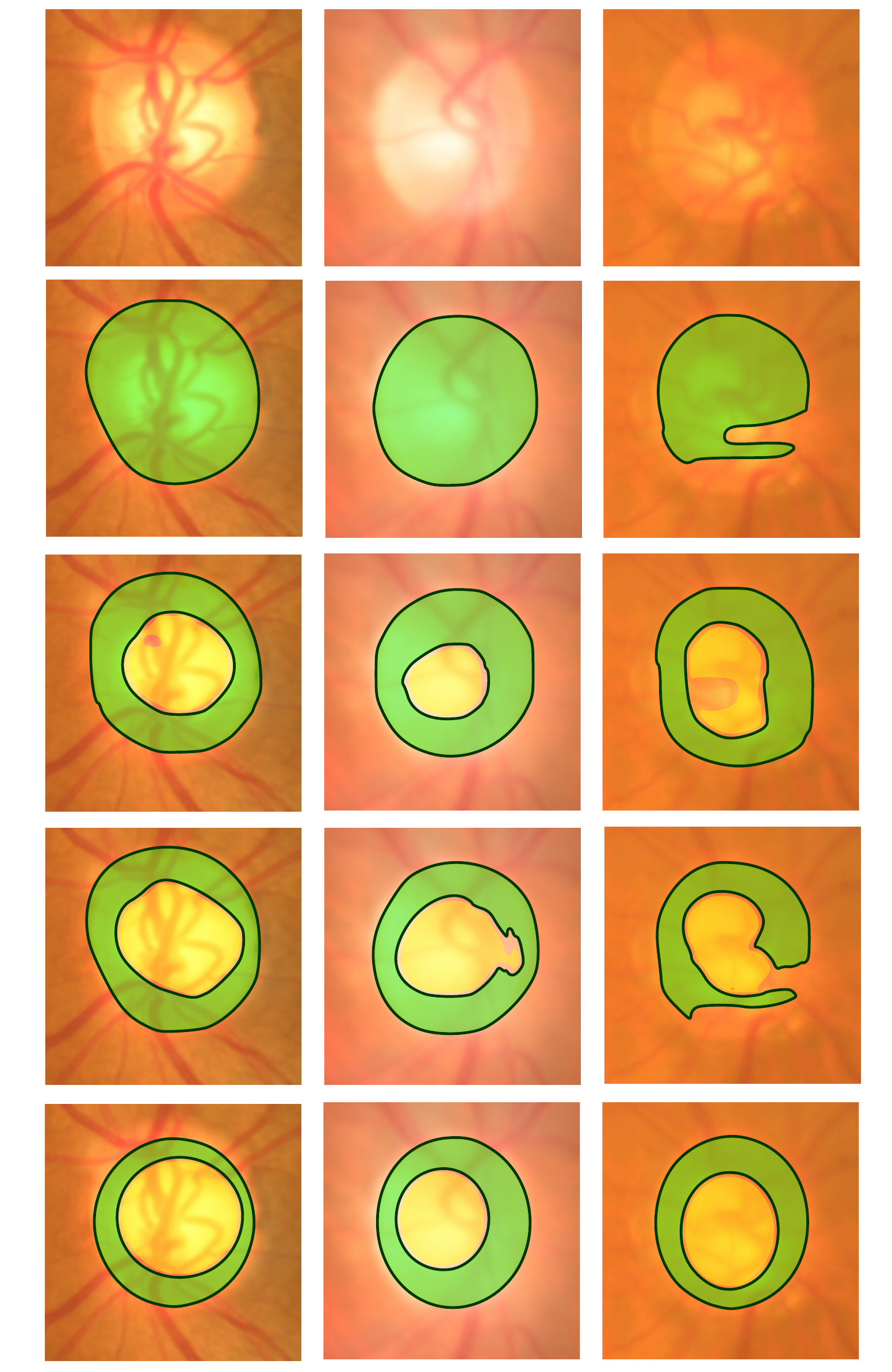}}%
    \vspace{0.2cm}
    \caption{Qualitative results obtained in (a) REFUGE, (b) DRISHTI, (c) RIM ONE v3, and (d) ORIGA for optic disc (green) and optic cup (yellow) segmentation. From top to bottom: detail of the image, coarse OD prediction from the TS model, OD/OC predictions obtained by the SS-model and the fine-grain stage of TS (OD $\rightarrow$ OD \& OC) and OD/OC ground truth segmentation.}
    \label{fig:qualitative-all}
\end{figure}

We performed an ablation study comparing all coarse-to-fine TS variants with respect to the SS one, all trained using the training set from our combined multi-dataset. 
Figure \ref{fig:ablation-study} depicts box-plots with the resulting distribution of Dice values obtained by each model both for OD and OC segmentation, while Figure~\ref{fig:qualitative-all} illustrates qualitative examples of the best, median and worst results in each of the test sets.

For the OD class (Figure~\ref{fig:ablation-study-od}), SS and TS approaches performed similarly in DRISHTI, REFUGE, RIM ONE v3 and ORIGA. In PAPILA, the SS model reported consistently better results than all TS variants, while in RIM ONE the model TS (OD - OD \& OC) reported similar but higher Dice values than the SS model. Hence, it is not possible to assume that the two-stage strategy uniformly contributes to improve OD results. This is also observed in Figure~\ref{fig:qualitative-all}, in which the overall shape of the OD segmented by both models is considerably similar.
On the other hand, coarse OD predictions (TS (OD - OC) in Figure~\ref{fig:ablation-study-od}) are in general more accurate than those obtained by the fine-grain stage (TS (OD - OD \& OC)). This result is coherent to the fact that the coarse stage dedicates all its capacity to identify the OD class, while the fine stage has to share it for segmenting OD/OC both at the same time.
Naturally, feeding the second stage with coarse OD prediction alleviates this issue, as we observed that TS (OD $\rightarrow$ OD \& OC) improved its fine grain results with respect to TS (OD - OD \& OC) in almost all datasets except RIM ONE. 
This stage achieves almost equal results than the coarse stage, meaning that this network does not introduce relevant corrections to its OD input but almost reproduces it at the output instead, as seen in Figure~\ref{fig:qualitative-all}.

As previously reported in REFUGE challenge~\cite{orlando2020refuge}, Dice values obtained for OC segmentation are in general lower than those achieved for the OD (Figure~\ref{fig:ablation-study}).
Similarly to what we observed in the OD analysis, TS variants outperformed the SS models in RIM ONE and (only slightly) in ORIGA for OC segmentation (Figure~\ref{fig:ablation-study-oc}), while SS obtained better results in DRISHTI, REFUGE and PAPILA. This again makes difficult to choose an optimal configuration. Qualitative observations in Figure~\ref{fig:qualitative-all} are equally inconclusive, with both alternatives showing sometimes either good or bad results.
Notice however that the SS model predicts both classes using the same number of parameters used by each stage in TS (OD - OC), meaning that the effective capacity of TS models is  higher. This might explain why TS models slightly improve results over the SS counterpart in some datasets.
Interestingly, supervising the fine grain stage with OD information showed a positive effect for OC segmentation in all test sets except for PAPILA (see improvements of TS (OD - OD \& OC) and (OD $\rightarrow$ OD \& OC) variants over (OD - OC) in Figure~\ref{fig:ablation-study-oc}).
This contribution seems to overweight the fact that the fine grain stage has to predict both classes at the same time with the same capacity.
Feeding the fine grain stage with the coarse OD prediction also improves OC results in DRISHTI, REFUGE and ORIGA, further emphasizing the relevance of counting with OD knowledge when segmenting the OC. Nevertheless, the overall increment in Dice values both in DRISHTI and REFUGE is not enough to make the TS (OD $\rightarrow$ OD \& OC) outperform the SS variant.

\begin{table}[]
\centering
\label{tab:dice}
\caption{Comparison with other state-of-the-art methods for OD/OC segmentation in terms of mean $\pm$ standard deviation Dice values. SS and TS models learned using only REFUGE or DRISHTI training sets are also included for comparison.}
\vspace{0.2cm}
\resizebox{\linewidth}{!}{ 
\begin{tabular}{l|cccc|cccc}
\hline
\multirow{2}{*}{Method} & \multicolumn{4}{c|}{Optic disc (OD) segmentation} & \multicolumn{4}{c}{Optic cup (OC) segmentation} \\ \cline{2-9}
 & REFUGE & DRISHTI & RIM ONE v3 & \multicolumn{1}{c|}{ORIGA} & REFUGE & DRISHTI & RIM ONE v3 & ORIGA \\ \hline
Al-Bandet \textit{et al}. (2018) \cite{al2018dense} & - & 0.949 & 0.904 & \multicolumn{1}{c|}{\textbf{0.965}} & - & 0.828 & 0.690 & 0.866 \\
Team CUHKMED (REFUGE, 2019) \cite{orlando2020refuge} & 0.960 & - & - & \multicolumn{1}{c|}{-} & 0.883 & - & - & - \\
Team Masker (REFUGE, 2019) \cite{orlando2020refuge} & 0.946 & - & - & \multicolumn{1}{c|}{-} & 0.884 & - & - & - \\
Team BUCT (REFUGE, 2019) \cite{orlando2020refuge} & 0.953 & - & - & \multicolumn{1}{c|}{-} & 0.873 & - & - & - \\
Wang \textit{et al}. (2019) \cite{wang2019patch} & 0.960 & 0.965 & 0.865 & \multicolumn{1}{c|}{-} & 0.883 & 0.858 & 0.787 & - \\
Shah \textit{et al}. (2019) \cite{shah2019dynamic}& - & 0.960 & 0.940 & \multicolumn{1}{c|}{-} & - & 0.890 & 0.820 & - \\
Tabassum \textit{et al}. (2020) \cite{tabassum2020cded} & - & 0.959 & \textbf{0.958} & \multicolumn{1}{c|}{-} & - & \textbf{0.924} & \textbf{0.862} & - \\
Liu \textit{et al}. (2021) \cite{liu2021joint} & 0.960 & \textbf{0.978} & - & \multicolumn{1}{c|}{-} & 0.890 & 0.912 & - & - \\
He \textit{et al}. (2022) \cite{he2022joined} & 0.954  & -  & - & - & 0.869 & - & - & - \\ \hline
SS-model (REFUGE)                                                                         & 0.943 $\pm$ 0.030 & 0.916 $\pm$ 0.077                  & 0.692 $\pm$ 0.255                  & 0.820 $\pm$ 0.243 & 0.845 $\pm$ 0.066 & 0.799 $\pm$ 0.127                  & 0.551 $\pm$ 0.262                  & 0.749 $\pm$ 0.238 \\
TS (OD $\rightarrow$ OD\&OC) (REFUGE)                                                     & 0.948 $\pm$ 0.027 & 0.934 $\pm$ 0.046                  & 0.670 $\pm$ 0.262                  & 0.788 $\pm$ 0.275 & 0.758 $\pm$ 0.142 & 0.804 $\pm$ 0.126                  & 0.483 $\pm$ 0.335                  & 0.691 $\pm$ 0.286 \\ \hline
SS-model (DRISHTI)                                                                        & 0.765 $\pm$ 0.245  & 0.940 $\pm$ 0.049                   & 0.830 $\pm$ 0.127                   & 0.678 $\pm$ 0.369  & 0.593 $\pm$ 0.246  & 0.801 $\pm$ 0.146                   & 0.536 $\pm$ 0.233                   & 0.557 $\pm$ 0.329  \\
TS (OD $\rightarrow$ OD\&OC) (DRISHTI)                                                    & 0.836 $\pm$ 0.167 & 0.948 $\pm$ 0.035                   & 0.850 $\pm$ 0.116                  & 0.762 $\pm$ 0.316 & 0.604 $\pm$ 0.261 & 0.801 $\pm$ 0.163                  & 0.579 $\pm$ 0.2401                 & 0.762 $\pm$ 0.326 \\ \hline
SS-model (multi-dataset)                                                                  & 0.954 $\pm$ 0.028 & 0.960 $\pm$ 0.017                  & 0.783 $\pm$ 0.215                  & 0.926 $\pm$ 0.055 & 0.868 $\pm$ 0.065 & 0.855 $\pm$ 0.119                  & 0.593 $\pm$ 0.252                  & 0.817 $\pm$ 0.124 \\
TS (OD $\rightarrow$ OD\&OC) (multi-dataset)                                              & 0.953 $\pm$ 0.033 & 0.955 $\pm$ 0.021                  & 0.804 $\pm$ 0.183                   & 0.920 $\pm$ 0.055 & 0.821 $\pm$ 0.110 & 0.835 $\pm$ 0.125                  & 0.643 $\pm$ 0.272                  & 0.831 $\pm$ 0.111 \\ \hline
\end{tabular}
}
\label{tab:dice}
\end{table}

Table \ref{tab:dice} includes a comparison with multiple state-of-the-art approaches in some of our test sets in terms of average Dice $\pm$ standard deviation. PAPILA results are not included here as we are the first to report numbers on this database. The table also includes results obtained by SS and TS when trained using only images from DRISHTI and from REFUGE. The TS-model corresponds to (OD $\rightarrow$ OD\&OC), which was the one exhibiting the most stable results among targets and databases in the ablation study. 
Increasing the amount and variability of training data seems to have a strong positive effect in both models. 
In particular, we observe that SS and TS trained using DRISHTI images (45 samples only) reported lower Dice values for almost all databases except for its own test set and RIM ONE v3. 
Models trained on REFUGE (400 images), on the other hand, are in general more accurate than those trained on DRISHTI, while those trained with our multi-dataset configuration (1120 images) outperformed them both.
We also observed that the TS model performs much better than the SS approach when only limited data is available. This can be seen when comparing the networks trained on DRISHTI, where TS usually reported higher Dice values than SS. When more training data is used such as in the REFUGE experiments, their differences are less obvious, with SS beating the TS approach in multiple test sets both for OD and OC segmentation.
Furthermore, when training data is also diverse as in e.g. our multi-dataset configuration, their average performance is almost equivalent, indicating that coarse-to-fine models might be more appropriate when less training samples are available.

We also observed that both SS and TS models trained in our multi-dataset outperform or are comparable with other existing approaches in these test sets. Some of these alternatives are coarse-to-fine networks that incorporate different domain-specific innovations~\cite{al2018dense,shah2019dynamic,wang2019patch,orlando2020refuge,liu2021joint}. 
In DRISHTI, results for OD segmentation are e.g outperformed by Liu \textit{et al.}~\cite{liu2021joint}, whose model is similar to our TS (OD - OD \& OC) architecture but implementing a different network design. 
When analyzing the OC class, our methods are most significantly outperformed by Tabassum \textit{et al.}~\cite{tabassum2020cded}, which introduces a novel architecture specifically crafted for computational efficiency during training. It should be pointed out, however, that their DRISHTI results are reported on a custom partition with 30\% of the original data, without following the pre-defined train/test split like ours.
Their average Dice values for OD/OC segmentation in RIM ONE v3 are also higher than the obtained using our SS and TS models, although they used 70\% of the database for training and again report results on the remaining 30\%. 
A similar decision was made by Shah \textit{et al.} \cite{shah2019dynamic} for training their coarse-to-fine method, which resembles our TS (OD - OD \& OC) but sharing the same encoder among the first and second stages and is trained in an end-to-end manner.
Instead, we used the entire RIM ONE v3 as a held-out test set. Therefore, our models did not see any images from it for training like in other studies \cite{shah2019dynamic,tabassum2020cded,wang2019patch}, which might explain why they performed much better. Al-Bandet et al \cite{al2018dense} also got better results in  RIM-ONE than us, probably due to training over cropped images instead of full fundus pictures, which resemble more the RIM ONE images.
Finally, in ORIGA dataset, Al-Bandet \textit{\textit{et al}.} \cite{al2018dense} reported higher segmentation results than ours. Their model consists of a U-shaped fully convolutional DenseNet combined with a postprocessing strategy that segments images that are manually cropped and rescaled. This method could then be seen as a two-stage approach in which the first part is handled by the user. Again, their model was trained on 70\% of the ORIGA data and evaluated in the remaining 30\%. On the contrary, our two models were trained on separate databases and evaluated in the entire ORIGA set.




\subsection{Application for glaucoma assessment}

\begin{figure}
    \centering
    
    \begin{subfigure}[]{0.19\textwidth}
      \includegraphics[width=\textwidth]{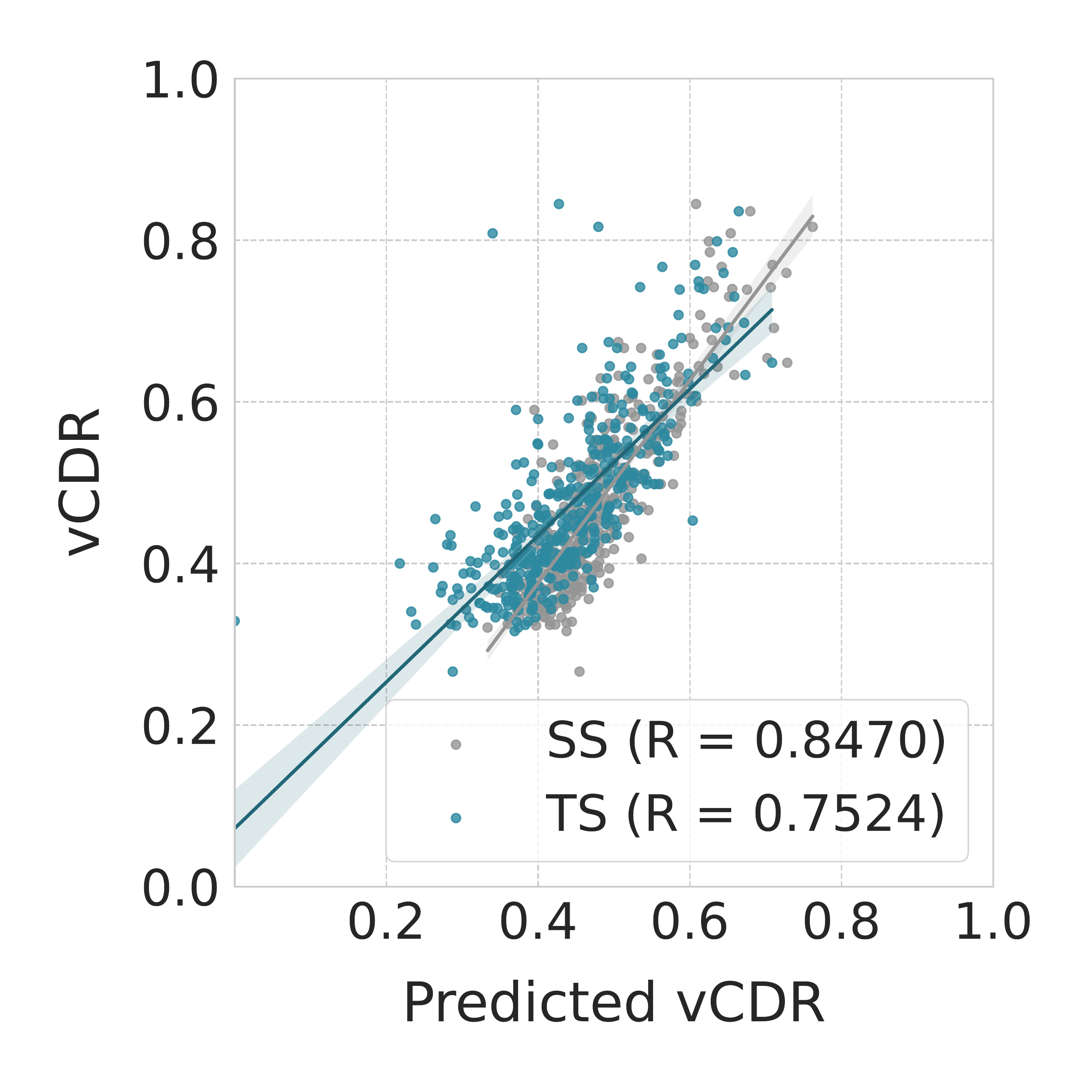}
      \includegraphics[width=\textwidth]{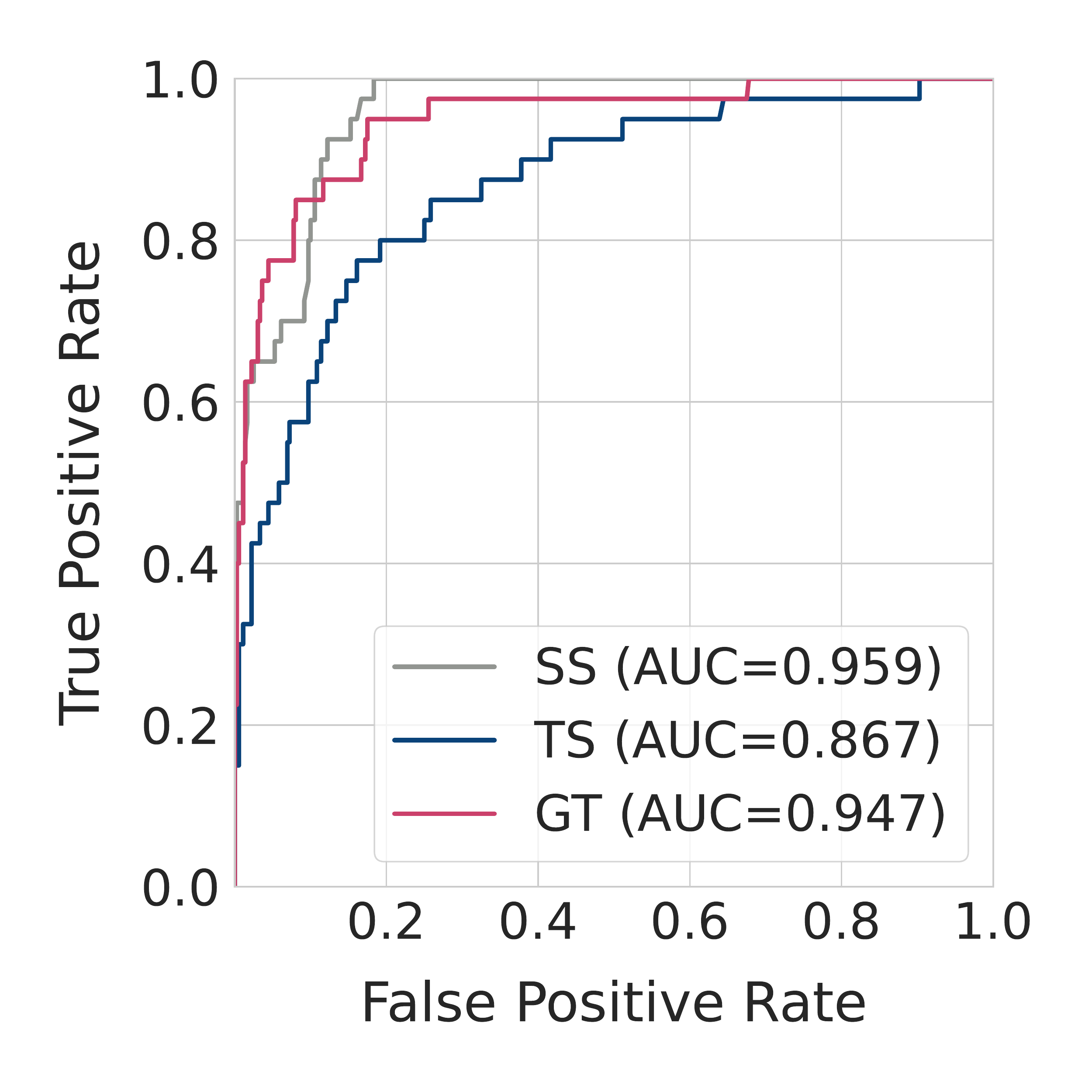}
      \caption{REFUGE}
      \label{fig:vcdrRefuge}
    \end{subfigure}
    \begin{subfigure}[]{0.19\textwidth}
      \includegraphics[width=\textwidth]{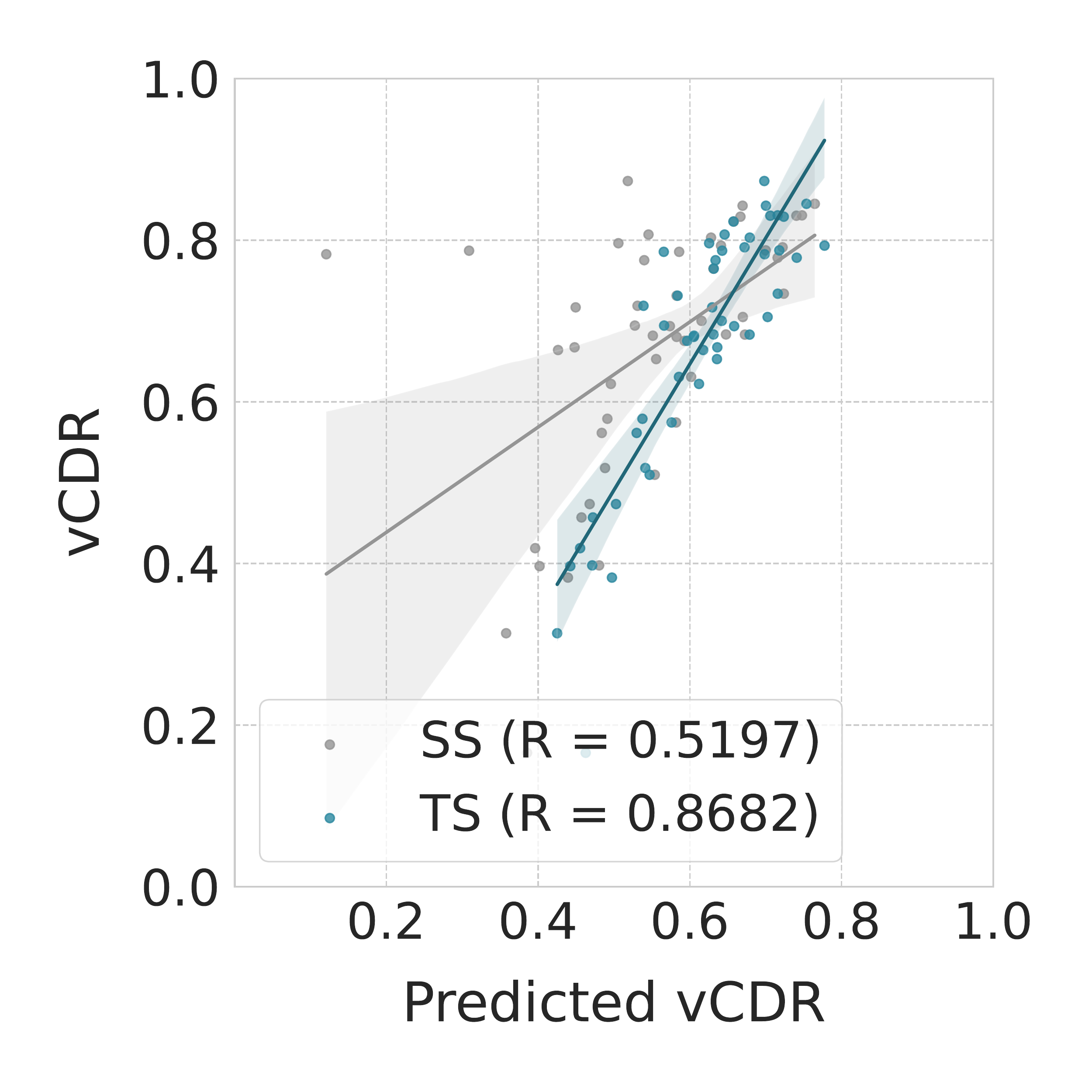}
      \includegraphics[width=\textwidth]{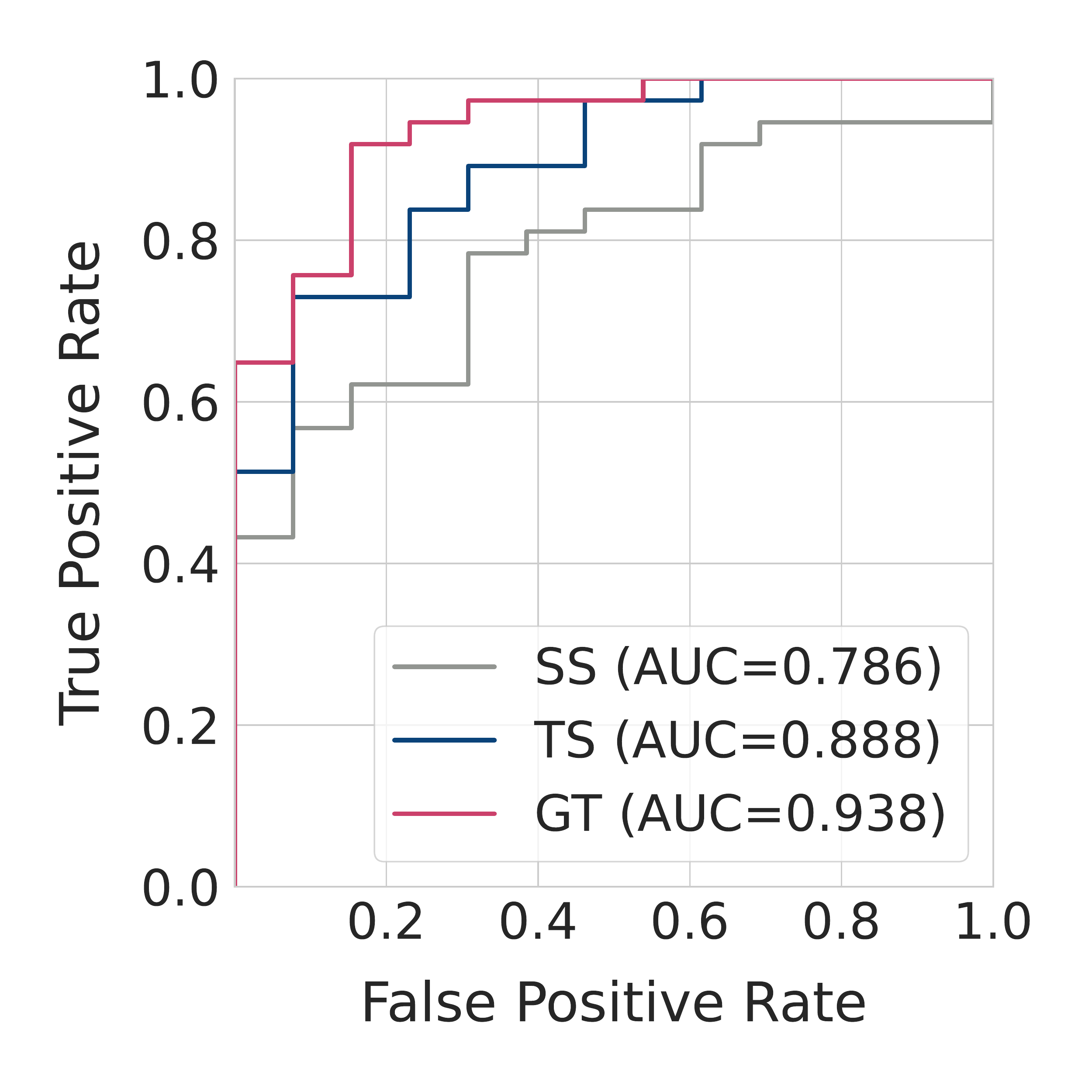}
      \caption{DRISHTI}
      \label{fig:vcdrDrishti}
    \end{subfigure}
    \begin{subfigure}[]{0.19\textwidth}
      \includegraphics[width=\textwidth]{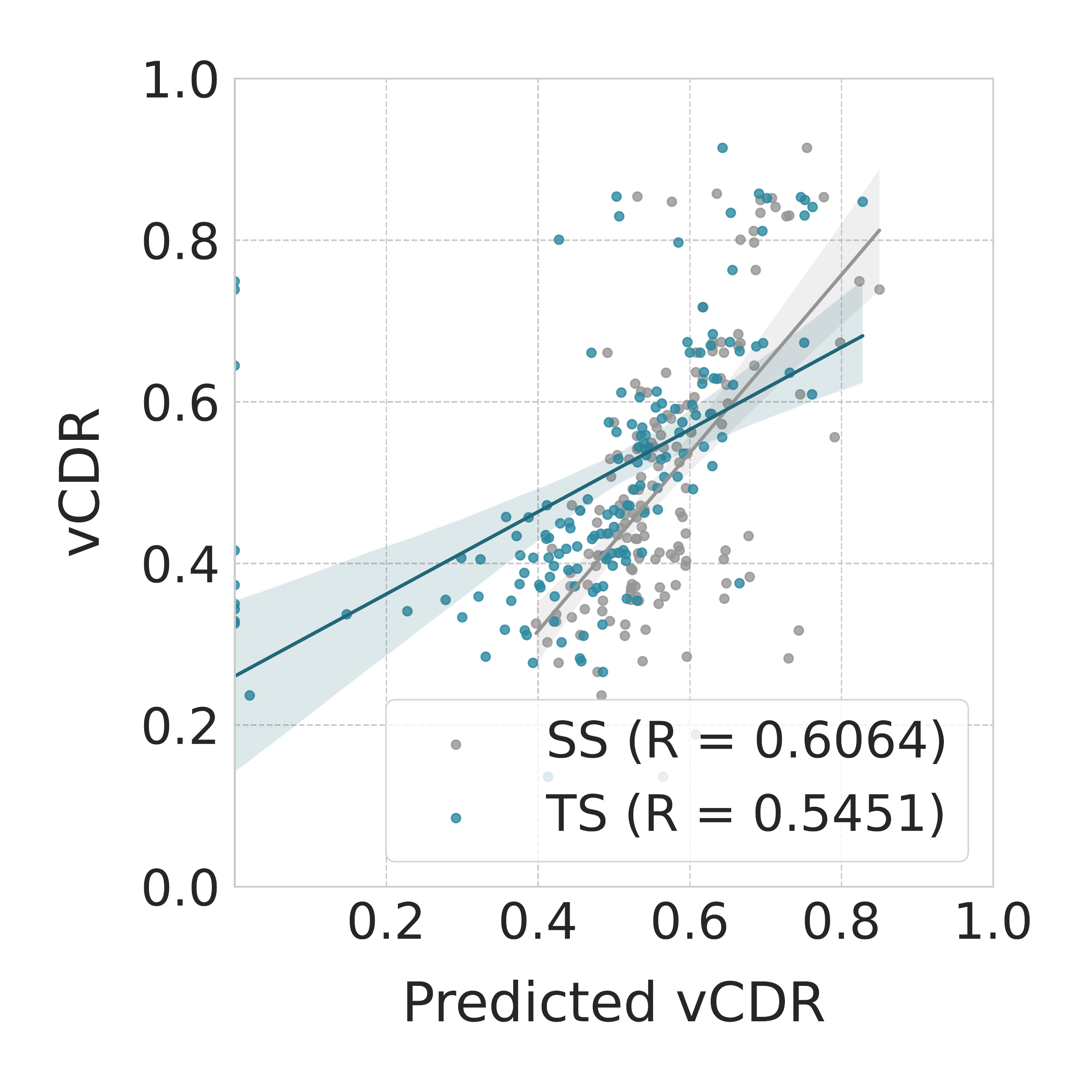}
      \includegraphics[width=\textwidth]{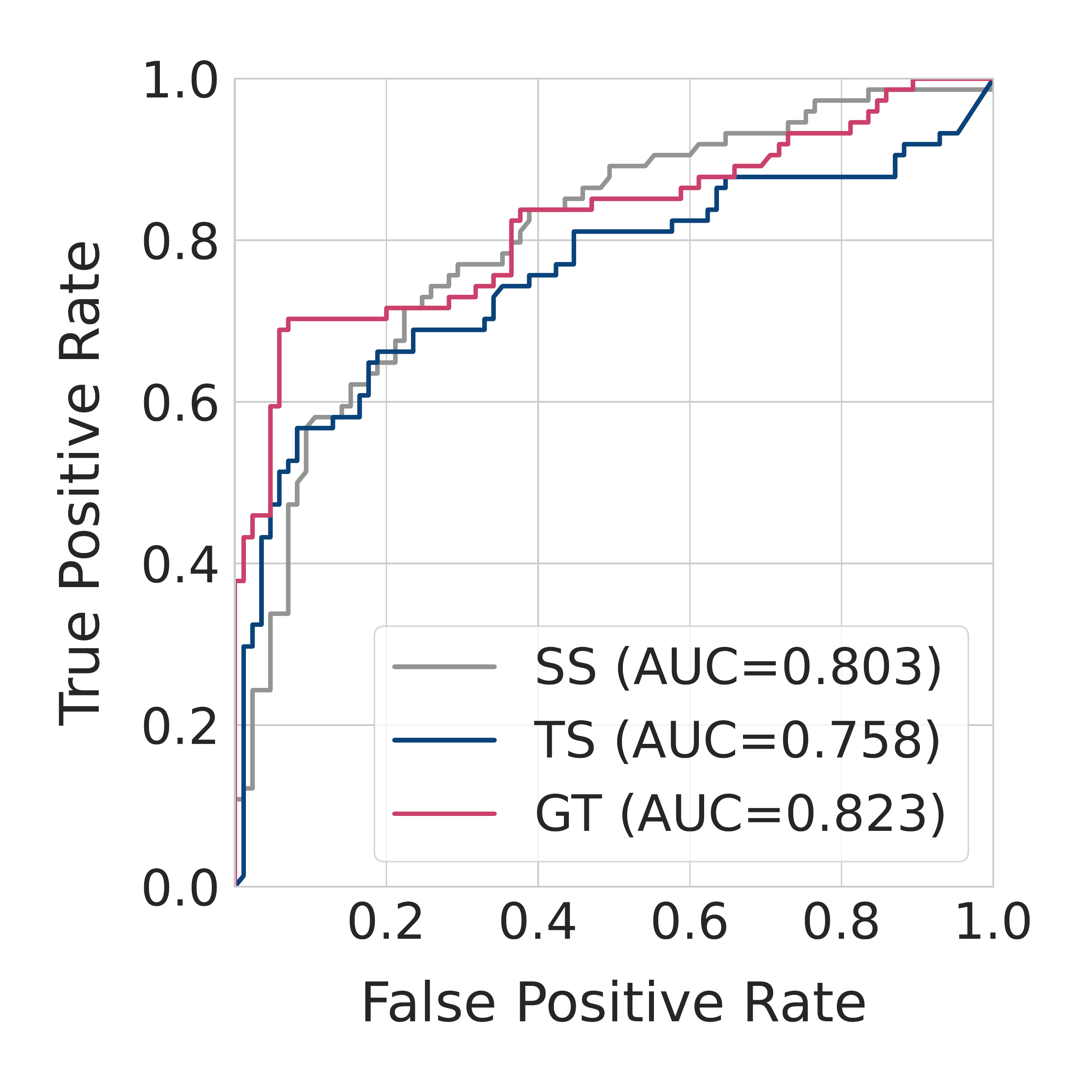}
      \caption{RIM ONE v3}
      \label{fig:vcdrRimone}
    \end{subfigure}
    \begin{subfigure}[]{0.19\textwidth}
      \includegraphics[width=\textwidth]{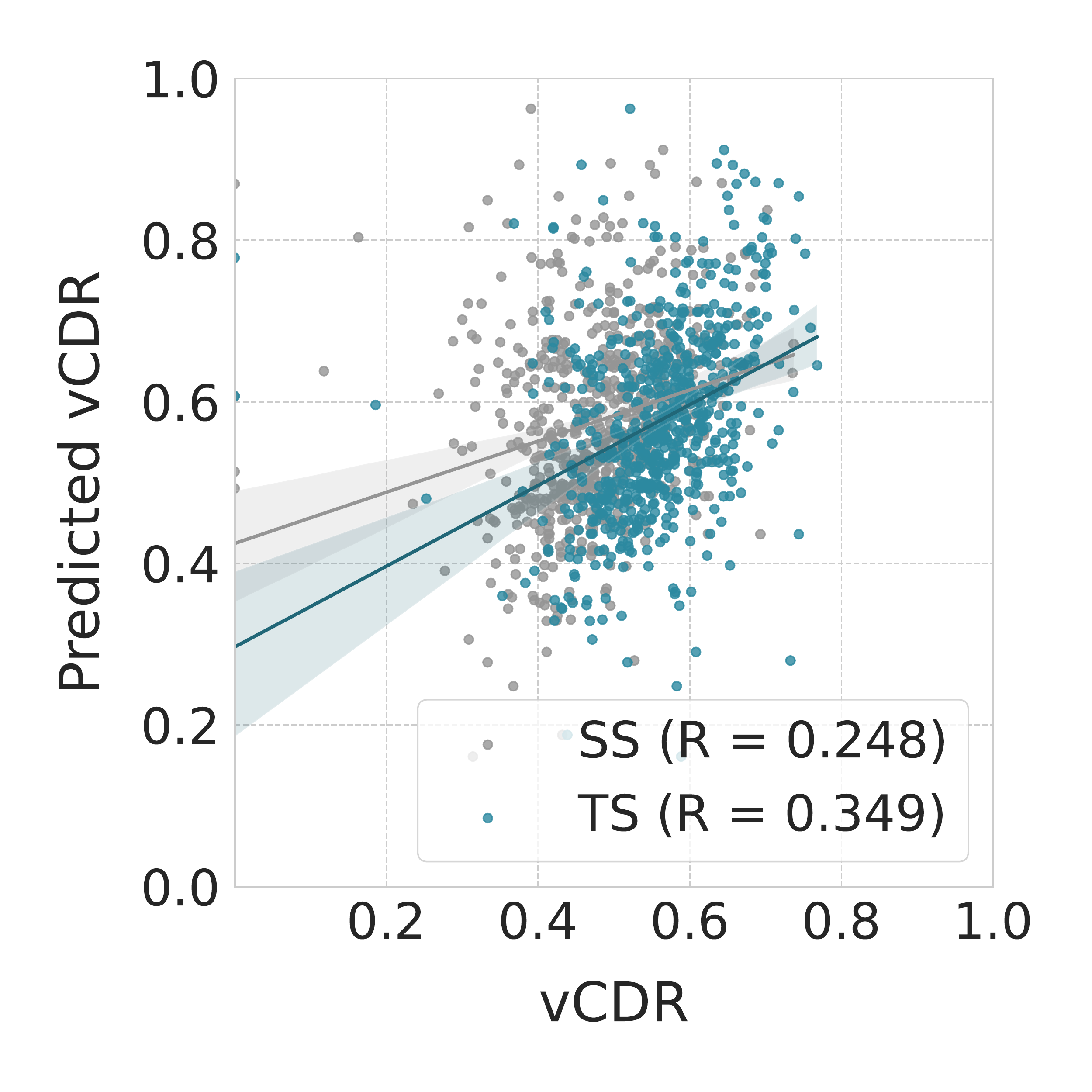}
      \includegraphics[width=\textwidth]{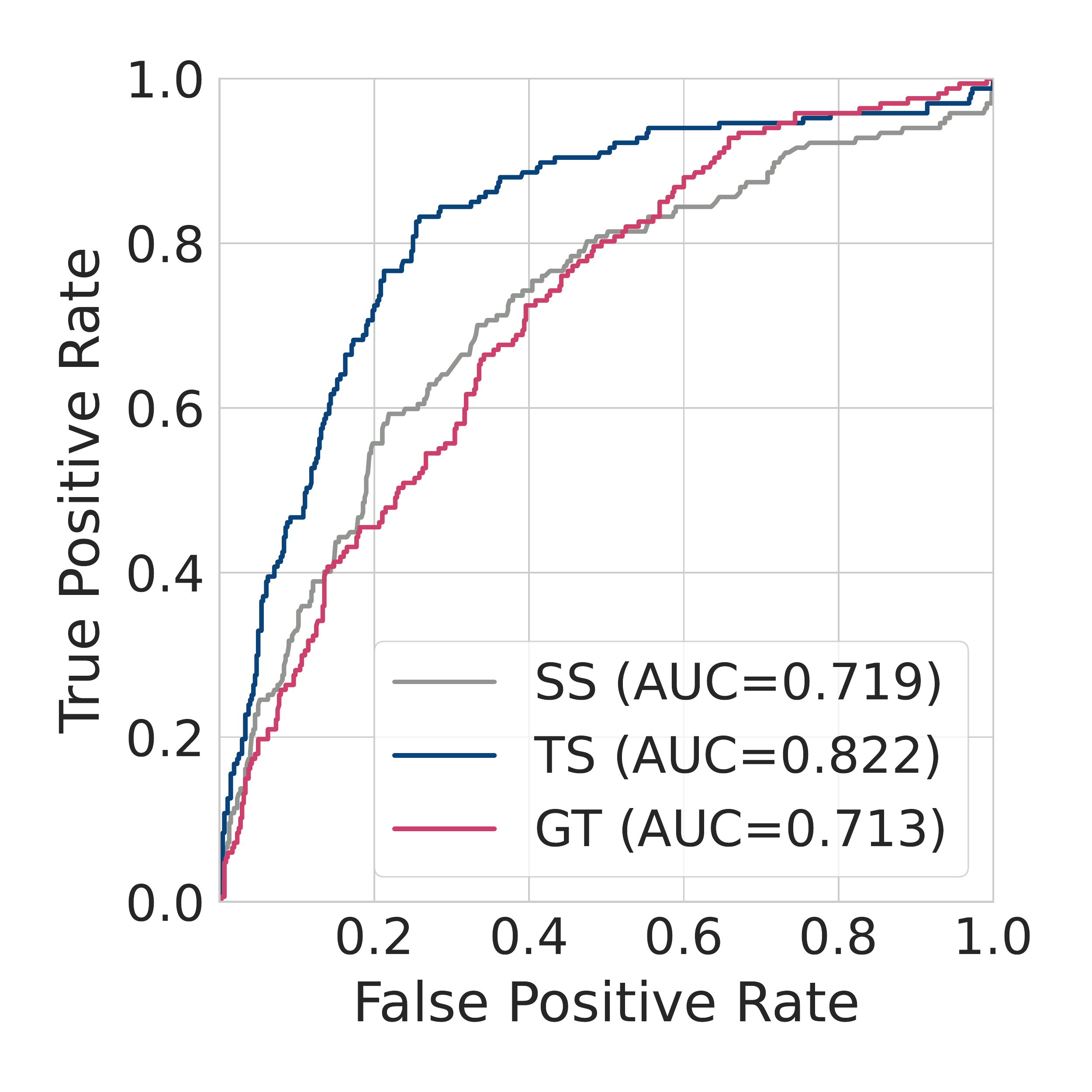}
      \caption{ORIGA}
      \label{fig:vcdrOriga}
    \end{subfigure}
    \begin{subfigure}[]{0.19\textwidth}
      \includegraphics[width=\textwidth]{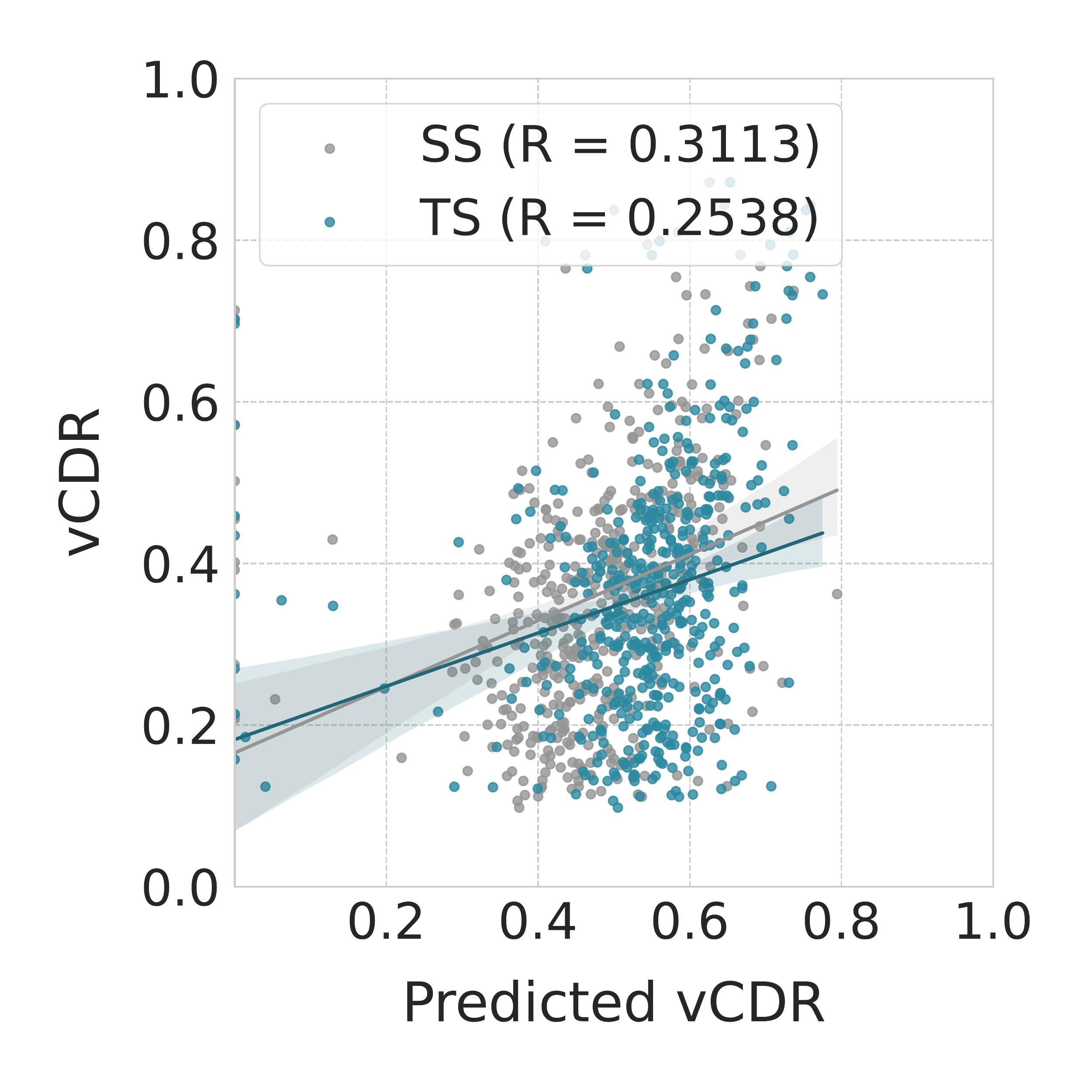}
      \includegraphics[width=\textwidth]{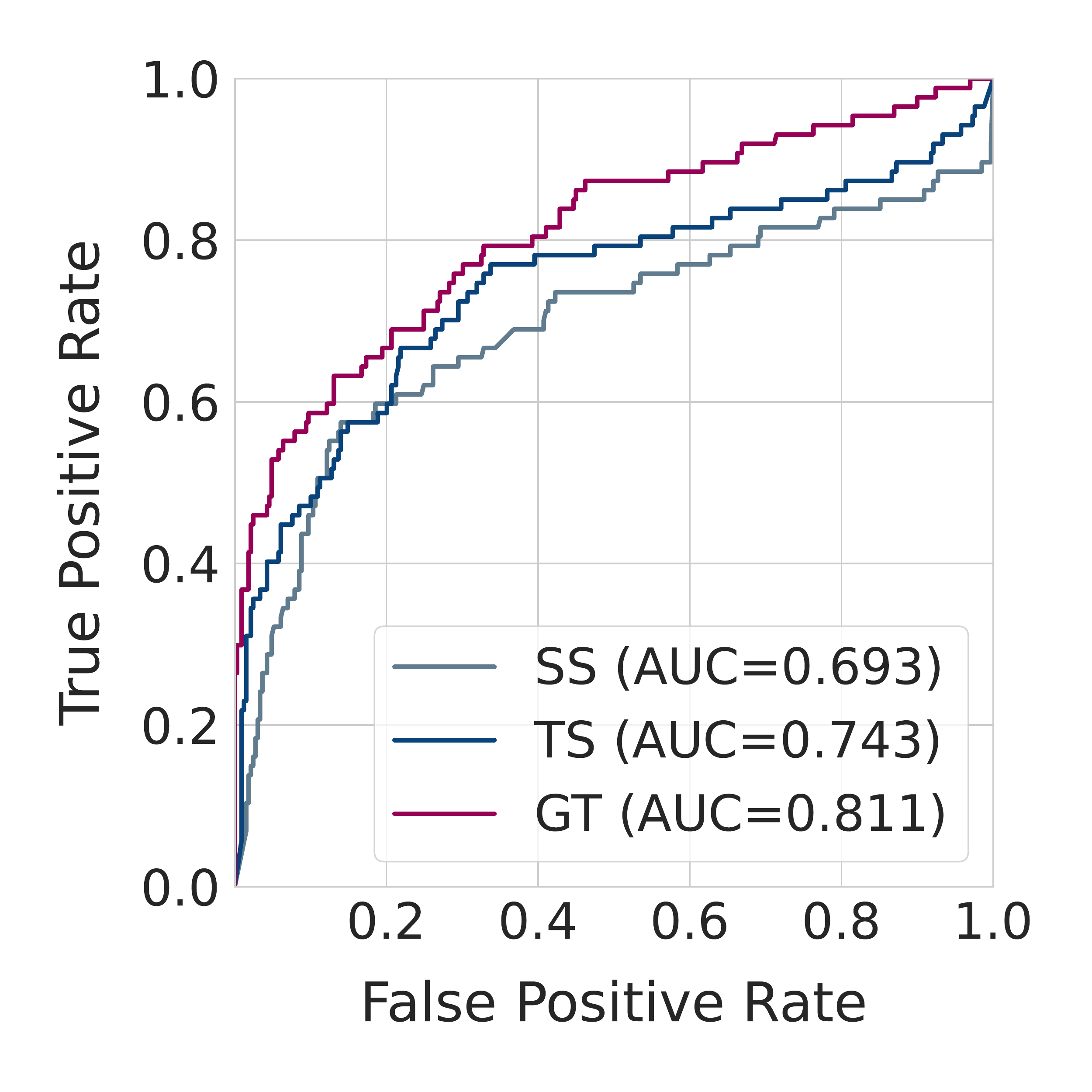}
      \caption{PAPILA}
      \label{fig:vcdrPapila}
    \end{subfigure}
    \vspace{0.2cm}
    \caption{Accuracy of vCDR predictions in (a) REFUGE, (b) DRISHTI, (c) RIM ONE v3, (d) ORIGA and (e) PAPILA. Top: scatter plots, regression lines and Pearson correlation coefficients $R$ between the vCDR predicted by the SS (gray) and TS (blue) models with respect to the vCDR computed from the ground truth segmentations. Bottom: ROC curves for glaucoma detection based on vCDR predicted by the SS (gray) and TS (blue) models, and calculated from the ground truth segmentations (red).}
    \label{fig:vcdr}
\end{figure}

We also evaluated the segmentation results in terms of their utility for computing the vCDR. Figure \ref{fig:vcdr} presents with scatter plots the agreement between vCDR values calculated from predictions obtained by the SS and TS models trained on the multi-dataset with respect to those obtained from the manual segmentations. Again, we used TS (OD $\rightarrow$ OD\&OC) due to its overall stability in the ablation study.
The highest $R$ values were observed in REFUGE and DRISHTI, while the lowest ones corresponds to those obtained in RIM ONE v3, ORIGA and PAPILA. This is consistent with the distribution of Dice values observed in Figure~\ref{fig:ablation-study}.
In REFUGE, RIM ONE v3 and PAPILA, the SS model reported better $R$ values than the TS approach. 
In PAPILA this might be linked with the fact that SS achieved better OD/OC results (Table~\ref{tab:dice}). In REFUGE, the performance for OD segmentation obtained by both models was similar, yet the SS model reported a higher average Dice for the OC (Table~\ref{tab:dice}).
On the other hand, TS beats SS R values in DRISHTI and ORIGA. While in ORIGA this could also be explained by differences in OC segmentation results, it is less straightforward to understand the behavior in RIM ONE and DRISHTI. One possible cause could be that the segmentations of SS and TS in each set, respectively, are better than their counterpart in their vertical appearance.

Figure~\ref{fig:vcdr} also includes ROC curves comparing the accuracy of the vCDR predictions for glaucoma detection in all our test sets.
In particular, TS obtained better glaucoma classification results in DRISHTI, ORIGA and PAPILA, while SS was better in REFUGE and RIM ONE v3. As observed in Section~\ref{subsec:results-segmentation}, again it is not clear if the coarse-to-fine model is actually better than the SS counterpart.



Notice also that high AUC values were obtained even in datasets where the correlations between the predicted vCDR and the ground truth measurements were not too high. This is more notorious in ORIGA (Figure~\ref{fig:vcdrOriga}), where the TS model reported an $R=0.349$ but an AUC $ = 0.822$ that is even higher than the one obtained using the manually measured vCDR.
Furthermore, ROC curves are coherent with the behavior observed in the scatter plots, indicating that models that were accurate in vCDR predictions are usually better for glaucoma detection. This holds true for all datasets except in PAPILA, where TS obtained lower $R$ values than SS but still reported a better AUC.
It is worth mentioning too that, despite the fact that OD/OC results are still not perfect, vCDR predictions derived from them perform equal or even better than manual measurements for glaucoma detection.

\begin{figure}
    \centering
    
   \includegraphics[width=0.8\textwidth]{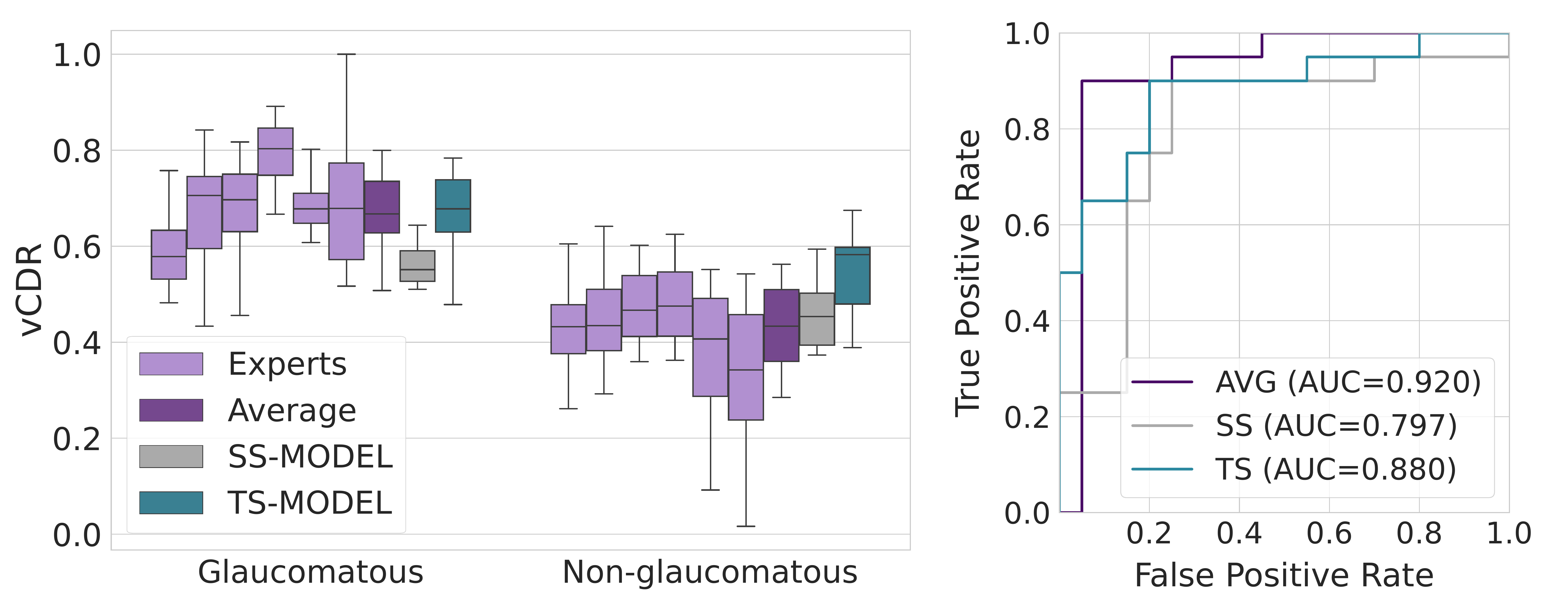}
   \caption{vCDR results in our multi-expert dataset derived from AIROGS. (Left) Distribution of vCDR measurements made by the six experts, their average and automated vCDR values calculated from our SS and TS models. (Right) ROC curves and their AUC values for glaucoma detection using the average vCDR of the experts (AVG) and the vCDR values estimated from our SS and TS models' outputs.} 
    \label{fig:vcdr_airogs}
\end{figure}

Finally, the left hand side of Figure \ref{fig:vcdr_airogs} presents the distribution of vCDR measurements made by the six experts in our AIROGS-derived dataset of fundus images, jointly with the vCDRs calculated from the SS and TS segmentations. We also computed the average of the vCDR responses of the experts and include it in the boxplot.
As previously mentioned in other studies~\cite{crowston2004effectvCDR,guo2019automatic}, a high inter-observer variability is seen among experts. These differences are more notorious in the glaucomatous subset, while in the non-glaucomatous set the median values are comparable one another.
Differences in vCDR predictions from SS and TS were also compelling in the glaucomatous set, where the TS model reported values similar to the average among experts and SS estimates were much lower. In the non-glaucomatous subset, on the other hand, we observed the inverse behavior, with SS obtaining similar results than the general trend, while the TS approach reported much higher values.
In Figure \ref{fig:vcdr_airogs} (right) we show the glaucoma detection ability of vCDR estimates obtained by averaging experts responses (AVG) or from segmentation outputs from our SS and TS models. The highest AUC was obtained using manual measurements, followed by the TS and SS estimates, respectively. This behavior aligns with the previous observation in Figure~\ref{fig:vcdr}, where we have seen comparable glaucoma classification results between manual and automatically predicted vCDR values.

\section{CONCLUSIONS}

In this paper we presented an in-depth empirical analysis of different coarse-to-fine approaches for OD/OC segmentation in fundus images. 
Surprisingly, we observed that single-stage (SS) models can perform equally well or even better than two-stage (TS) approaches if trained using large sets of diverse images. This seems to indicate that two-stage designs are more appropriate when training data is scarce.
Our ablation study also showed the importance of providing OD supervision to the fine stage, as the network trained solely to predict the OC obtained worst Dice values than the one that segments both OD/OC at the same time.
Results also showed that using the output of the coarse stage to augment the input of the fine stage improves results over the commonly adopted strategy of simply using the image cropped around the optic nerve head as input.
To the best of our knowledge, incorporating the coarse segmentation to the fine stage input was never used before.
Nevertheless, we observed that the second stage in this case does not introduce significant modifications to the coarse mask. Future studies should focus in enforcing the fine stage to better exploit this prior knowledge, e.g. by using multiscale gatings~\cite{fu2018joint}
Another interesting result is that, even without any domain-specific methodological innovation or custom postprocessing, both our SS and TS approaches are close to the top ranked models in REFUGE. 
Furthermore, they both reported similar average Dice values than other more complex methods in DRISHTI. 
We believe that incorporating other mechanisms to encourage shape consistency e.g. using adversarial or contour-based losses might aid to further improve these segmentations.
Nevertheless, these masks were accurate enough to obtain vCDR estimates that were able to achieve glaucoma classification results comparable to those obtained from manual measurements.

\acknowledgments 

This study was partially funded by PICT 2019-00070 and PICT startup 2021-00023 granted by Agencia I+D+i (Ministerio de Ciencia, Tecnología e Innovación de la Nación, Argentina) and by PIP 2021-2023 11220200102472CO from CONICET (Argentina). We also thank NVIDIA Corporation for granting cloud-based GPU computing hours for this project as part of a NVIDIA Hardware Grant.


\bibliography{report} 

\begin{thebibliography}{10}

\bibitem{orlando2020refuge}
Orlando, J.~I. et~al., ``{REFUGE} challenge: A unified framework for evaluating
  automated methods for glaucoma assessment from fundus photographs,'' {\em
  Med. Image Anal.}~{\bf 59},  101570 (2020).

\bibitem{kovalyk2022papila}
Kovalyk, O. et~al., ``{PAPILA}: Dataset with fundus images and clinical data of
  both eyes of the same patient for glaucoma assessment,'' {\em Sci. Data}~{\bf
  9}(1),  1--12 (2022).

\bibitem{phene2019deep}
Phene, S. et~al., ``Deep learning and glaucoma specialists: the relative
  importance of optic disc features to predict glaucoma referral in fundus
  photographs,'' {\em Ophthalmology}~{\bf 126}(12),  1627--1639 (2019).

\bibitem{li2021applications}
Li, T. et~al., ``Applications of deep learning in fundus images: A review,''
  {\em Med. Image Anal.}~{\bf 69},  101971 (2021).

\bibitem{crowston2004effectvCDR}
Crowston, J. et~al., ``The effect of optic disc diameter on vertical cup to
  disc ratio percentiles in a population based cohort: the {Blue Mountains Eye
  Study},'' {\em BJO}~{\bf 88}(6),  766--770 (2004).

\bibitem{guo2019automatic}
Guo, J. et~al., ``Automatic determination of vertical cup-to-disc ratio in
  retinal fundus images for glaucoma screening,'' {\em IEEE Access}~{\bf 7},
  8527--8541 (2019).

\bibitem{liu2021joint}
Liu, B., Pan, D., and Song, H., ``Joint optic disc and cup segmentation based
  on densely connected depthwise separable convolution deep network,'' {\em BMC
  Med.}~{\bf 21}(1),  1--12 (2021).

\bibitem{shah2019dynamic}
Shah, S., Kasukurthi, N., and Pande, H., ``Dynamic region proposal networks for
  semantic segmentation in automated glaucoma screening,'' in [{\em ISBI
  2019}{\nolinebreak\hspace{0.1em}]},   578--582, IEEE (2019).

\bibitem{wang2019patch}
Wang, S. et~al., ``Patch-based output space adversarial learning for joint
  optic disc and cup segmentation,'' {\em IEEE Trans. Med. Imaging}~{\bf
  38}(11),  2485--2495 (2019).

\bibitem{tabassum2020cded}
Tabassum, M. et~al., ``{CDED-Net}: Joint segmentation of optic disc and optic
  cup for glaucoma screening,'' {\em IEEE Access}~{\bf 8},  102733--102747
  (2020).

\bibitem{al2018dense}
Al-Bander, B., , et~al., ``Dense fully convolutional segmentation of the optic
  disc and cup in colour fundus for glaucoma diagnosis,'' {\em Symmetry}~{\bf
  10}(4),  87 (2018).

\bibitem{fang2022refuge2}
Fang, H. et~al., ``{REFUGE2} challenge: Treasure for multi-domain learning in
  glaucoma assessment,'' {\em arXiv preprint arXiv:2202.08994}  (2022).

\bibitem{almazroa2018retinal}
Almazroa, A. et~al., ``Retinal fundus images for glaucoma analysis: the {RIGA}
  dataset,'' in [{\em Medical Imaging 2018: Imaging Informatics for Healthcare,
  Research, and Applications}{\nolinebreak\hspace{0.1em}]},   {\bf 10579},
  55--62, SPIE (2018).

\bibitem{zhang2010origa}
Zhang, Z. et~al., ``Origa-light: An online retinal fundus image database for
  glaucoma analysis and research,'' in [{\em EMBC
  2010}{\nolinebreak\hspace{0.1em}]},   3065--3068, IEEE (2010).

\bibitem{fumero2011rim}
Fumero, F. et~al., ``{RIM-ONE}: An open retinal image database for optic nerve
  evaluation,'' in [{\em CMBS 2014}{\nolinebreak\hspace{0.1em}]},   1--6, IEEE
  (2011).

\bibitem{sivaswamy2014drishti}
Sivaswamy, J. et~al., ``{Drishti-GS}: Retinal image dataset for optic nerve
  head ({ONH}) segmentation,'' in [{\em ISBI
  2014}{\nolinebreak\hspace{0.1em}]},   53--56, IEEE (2014).

\bibitem{porwal2018indian}
Porwal, P. et~al., ``Indian diabetic retinopathy image dataset ({IDRiD}): a
  database for diabetic retinopathy screening research,'' {\em Data}~{\bf
  3}(3),  25 (2018).

\bibitem{alawad2022machine}
Alawad, M. et~al., ``Machine learning and deep learning techniques for optic
  disc and cup segmentation--a review,'' {\em Clinical Ophthalmology (Auckland,
  NZ)}~{\bf 16},  747 (2022).

\bibitem{he2022joined}
He, H., Lin, L., Cai, Z., and Tang, X., ``{JOINED}: Prior guided multi-task
  learning for joint optic disc/cup segmentation and fovea detection,'' in
  [{\em PMLR}{\nolinebreak\hspace{0.1em}]},  (2022).

\bibitem{fu2018joint}
Fu, H. et~al., ``Joint optic disc and cup segmentation based on multi-label
  deep network and polar transformation,'' {\em IEEE TMI}~{\bf 37}(7),
  1597--1605 (2018).

\end{thebibliography}
\bibliographystyle{spiebib} 

\end{document}